\documentclass[]{bytedance_seed}

\usepackage{minitoc}

\usepackage{solarized-light}
\usepackage{booktabs}
\usepackage{multirow}
\usepackage{graphicx}
\usepackage{array}
\usepackage{makecell}
\usepackage{framed}
\usepackage{hyperref}
\usepackage{amsmath}  
\usepackage{amssymb}
\usepackage[colorinlistoftodos]{todonotes}
\usepackage{longtable}
\usepackage{hhline}
\usepackage{fancyvrb}
\usepackage{fvextra}
\usepackage{CJKutf8}
\usepackage{multicol}
\usepackage{pifont}
\usepackage{cleveref}
\usepackage{tablefootnote}
\usepackage{threeparttable}
\usepackage{tabularx}
\usepackage{mdframed}
\usepackage{subcaption}
\usepackage[usestackEOL]{stackengine}
\usepackage[numbers]{natbib}
\newcommand{\commentout}[1]{}
\renewcommand{\paragraph}[1]{\noindent\textbf{#1.}\hspace*{1em}}
\usepackage{enumitem}
\setlist[itemize]{leftmargin=15pt}
\usepackage{xcolor}
\usepackage{colortbl}

\RequirePackage{xspace}
\makeatletter
\DeclareRobustCommand\onedot{\futurelet\@let@token\@onedot}
\def\@onedot{\ifx\@let@token.\else.\null\fi\xspace}

\makeatother

\newcommand{\uitars}{UI-TARS\xspace}

\title{UI-TARS-2 Technical Report: Advancing GUI Agent \\ with Multi-Turn Reinforcement Learning}

\author{ByteDance Seed}

\contribution{See \hyperref[sec:contributions]{Contributions} section for a full author list.}

\abstract{
The development of autonomous agents for graphical user interfaces (GUIs) presents major challenges in artificial intelligence.  
While recent advances in native agent models have shown promise by unifying perception, reasoning, action, and memory through end-to-end learning, open problems remain in data scalability, multi-turn reinforcement learning (RL), the limitations of GUI-only operation, and environment stability.  
In this technical report, we present UI-TARS-2, a native GUI-centered agent model that addresses these challenges through a systematic training methodology: a data flywheel for scalable data generation, a stabilized multi-turn RL framework, a hybrid GUI environment that integrates file systems and terminals, and a unified sandbox platform for large-scale rollouts.  
Empirical evaluation demonstrates that UI-TARS-2 achieves significant improvements over its predecessor UI-TARS-1.5.  
On GUI benchmarks, it reaches 88.2 on Online-Mind2Web, 47.5 on OSWorld, 50.6 on WindowsAgentArena, and 73.3 on AndroidWorld, outperforming strong baselines such as Claude and OpenAI agents.  
In game environments, it attains a mean normalized score of 59.8 across a 15-game suite-roughly 60\% of human-level performance-and remains competitive with frontier proprietary models (e.g., OpenAI o3) on LMGame-Bench.  
Additionally, the model can generalize to long-horizon information-seeking tasks and software engineering benchmarks, highlighting its robustness across diverse agent tasks. 
Detailed analyses of training dynamics further provide insights into achieving stability and efficiency in large-scale agent RL.  
These results underscore UI-TARS-2's potential to advance the state of GUI agents and exhibit strong generalization to real-world interactive scenarios.
\footnotetext[1]{Project: \url{https://github.com/bytedance/ui-tars}, \url{https://github.com/bytedance/UI-TARS-desktop}}  
\footnotetext[2]{Demo: \url{https://seed-tars.com/showcase/ui-tars-2}}
}
\correspondence{\email{yujia.qin@bytedance.com};
\email{shiguang.sg@bytedance.com}}
\begin{document}

\maketitle

\newpage
\tableofcontents
\newpage

\section{Introduction}
The development of agents that can operate seamlessly within graphical user interfaces (GUIs) has emerged as a central challenge in artificial intelligence~\citep{zhang2024large,nguyen2024gui,wang2024gui,hu2025agents,tang2025survey}. Traditional approaches typically adopt modular pipelines with separately engineered components for perception, planning, memory, and action~\citep{liu2025advances}. While such design-driven systems enable rapid progress in specific domains, they rely heavily on expert heuristics and task-specific rules, leaving them brittle and difficult to scale. Recent work on \textit{native agent models}~\citep{qin2025ui} shifts toward data-driven, end-to-end learning, where perception, reasoning, and control are unified within a single policy, offering a more scalable and adaptive path for GUI agents.

Despite recent progress, the development of robust GUI agents still faces several open challenges.  
\textbf{(1) Data scarcity.} While large-scale pre-training and reinforcement learning have proven effective in reasoning and chat domains~\citep{openaichatgptblog,jaech2024openai}, scalable strategies for long-horizon GUI learning remain unclear. Unlike text or code corpora, large-scale trajectories that capture detailed reasoning, actions, environment states, and feedback are extremely costly to collect.
\textbf{(2) Scalable multi-turn RL.} RL in interactive environments is notoriously difficult: rewards are often sparse or delayed, optimization can be unstable, and credit assignment across long sequences of actions remains challenging. These issues hinder scaling beyond short-horizon demonstrations and make it hard to achieve stable improvements in complex tasks. 
\textbf{(3) Limitations of GUI-only operation.} Pure GUI interaction is often insufficient for realistic workflows. Many tasks—such as data processing, software development, or system administration—are more naturally handled through file systems, terminals, or external tools rather than by simulating mouse clicks and keystrokes. Thus, advancing GUI agents requires environments that allow graphical actions to interoperate seamlessly with other resources, broadening the scope of tasks they can effectively solve.  
\textbf{(4) Environment scalability and stability.} Even with richer interaction capabilities, deploying large-scale RL environments remains an engineering bottleneck. Rollouts must be reproducible, fault-tolerant, and capable of supporting millions of interactive episodes across browsers, VMs, and simulators. In practice, such environments are fragile, resource-intensive, and prone to crashes, making stable large-scale training particularly challenging.

To address these challenges, we introduce a systematic methodology built on four pillars. First, to mitigate data scarcity, we design a scalable \textbf{Data Flywheel} that co-evolves the model and its training corpus through continual pre-training, supervised fine-tuning, rejection sampling, and multi-turn RL. This framework supplies a steady stream of diverse, high-quality trajectories and ensures that both the model and the data improve iteratively in a self-reinforcing cycle.  
Second, to overcome the difficulties of \textbf{scalable multi-turn RL}, we design a training framework that stabilizes optimization in long-horizon settings. 
This includes asynchronous rollouts with stateful environments to preserve context, streaming updates to avoid bottlenecks from long-tail trajectories, and enhanced proximal policy optimization~\citep{schulman2017proximal} with reward shaping, adaptive advantage estimation, and value pretraining. 
Third, to move beyond the limitations of pure GUI interaction, we construct a \textbf{hybrid GUI-centered environment} that augments on-screen actions with access to complementary resources such as file systems, terminals, and other external tools, enabling agents to solve a broader spectrum of realistic workflows.  
Fourth, to support large-scale training and evaluation, we build a \textbf{unified sandbox platform} capable of orchestrating heterogeneous environments—ranging from cloud VMs for GUI interaction to browser-based sandboxes for games—under a consistent API. The platform is engineered for reproducibility, stability, and high throughput, making it possible to run millions of interactive rollouts reliably.

Empirical evaluation shows that UI-TARS-2 delivers significant improvements over UI-TARS-1.5~\citep{seed2025uitars15}, achieving strong results in both GUI-based interaction and game environments.  
On GUI benchmarks, the model reaches 88.2 on Online-Mind2Web~\citep{xue2025illusionprogressassessingcurrent}, 47.5 on OSWorld~\citep{xie2024osworld}, 50.6 on WindowsAgentArena~\citep{bonatti2024windows}, and 73.3 on AndroidWorld~\citep{rawles2024androidworld}, representing clear gains over the previous generation and outperforming strong baselines such as Claude and OpenAI agents in multiple cases.  
In game environments, UI-TARS-2 attains a mean normalized score of 59.8 across a 15-game suite—roughly 60\% of human-level performance—and surpasses strong baselines such as OpenAI CUA and Claude Computer Use by factors of 2.4× and 2.8×, respectively.  On LMGame-Bench~\citep{hu2025lmgamebenchgoodllmsplaying}, UI-TARS-2 remains competitive with frontier proprietary models, further highlighting its robustness in long-horizon game reasoning. 
Beyond GUI and games, we extend the agent's capabilities through GUI-SDK, enabling integration with system-level resources such as terminals and external tools.  
With this extension, UI-TARS-2 demonstrates strong performance on long-horizon information-seeking benchmarks (e.g., 29.6 on BrowseComp~\citep{wei2025browsecomp}) and competitive results on software engineering tasks (45.3 on Terminal Bench~\citep{tbench_2025}, 68.7 on SWE-Bench Verified~\citep{jimenez2023swebench}).  
These results suggest that the training methodology developed for GUI agents—particularly multi-turn RL optimization and scalable rollout infrastructure—transfers effectively to other interactive domains, broadening the agent’s applicability.  
In addition, our detailed analyses of training dynamics, interaction scaling, and hybrid training strategies provide practical insights into achieving stability and efficiency in large-scale agent reinforcement learning.  
Taken together, these findings establish UI-TARS-2 as a robust GUI-centered agent that not only advances the state of GUI interaction but also generalizes effectively to diverse real-world environments.  
\section{UI-TARS-2}
\label{sec:methodology}
\begin{figure*}[t]
    \centering
\includegraphics[width=0.95\linewidth]{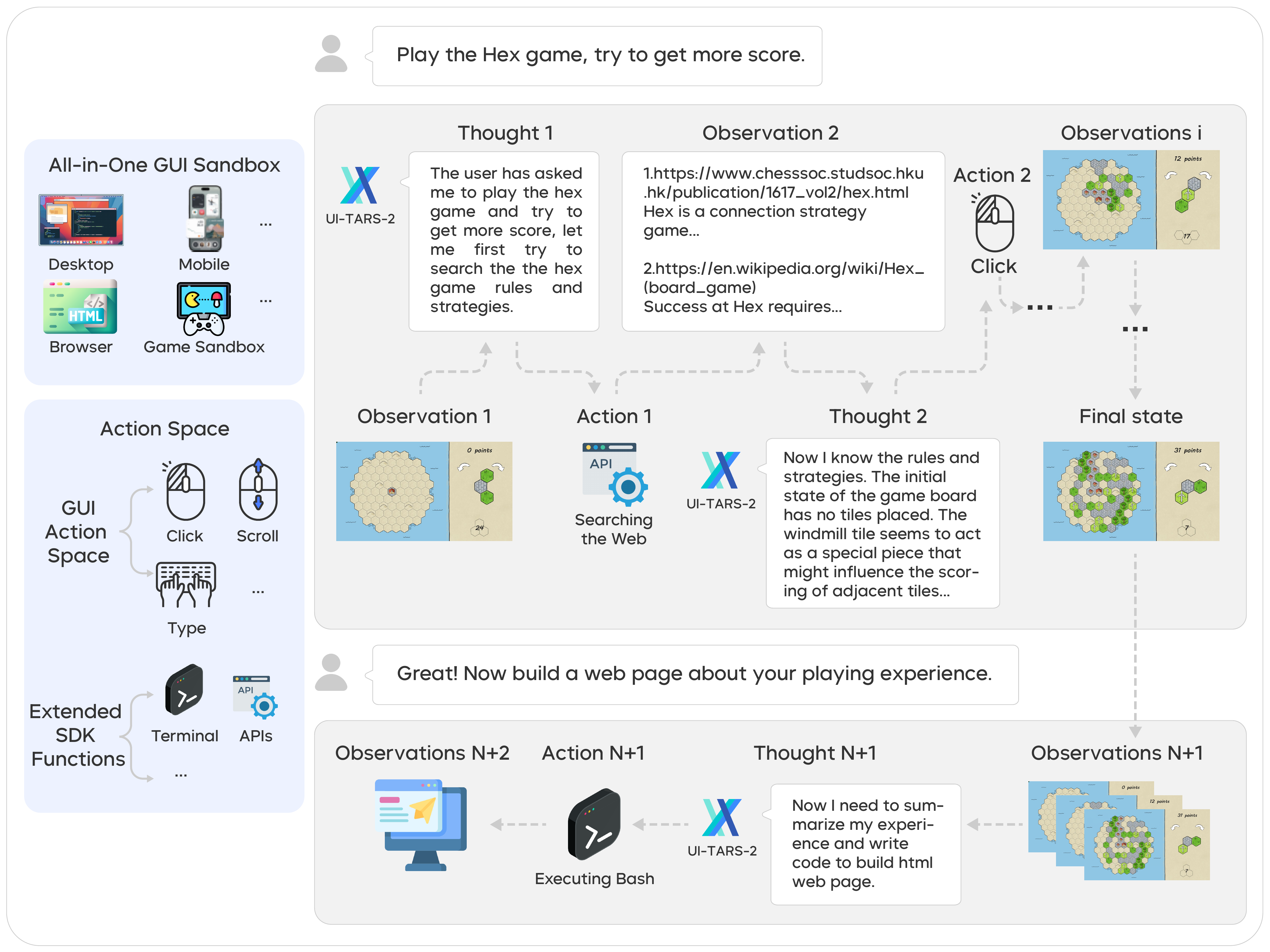}
    \caption{
    A demo trajectory of UI-TARS-2.
    }
    \label{fig:omni}
\end{figure*}

This section introduces the methodology of UI-TARS-2, a unified framework for building advanced GUI-centered agents. We illustrate a demo trajectory of UI-TARS-2 in Figure~\ref{fig:omni}.
Our approach integrates multiple components, including formal agent formulation, all-in-one sandbox environments, a data flywheel pipeline, multi-turn reinforcement learning, and parameter interpolation across vertical agents.

\subsection{Formulation}
We adopt a native agent perspective~\citep{qin2025ui}, where an agent is modeled as a parameterized policy that maps historical context, memory states, and the current environment into behavioral outputs. At timestep $t$, the agent follows the ReAct paradigm~\citep{yao2023react}, which interleaves reasoning, action, and observation in a structured loop:  
\begin{itemize}
    \item \textbf{Reasoning ($t_t$)}: internal cognitive processing, including context analysis, memory recall, planning, and self-reflection.  
    \item \textbf{Action ($a_t$)}: external interaction, such as GUI manipulation, system commands, or tool invocation.  
    \item \textbf{Observation ($o_t$)}: feedback from the environment used to update the agent's state.  
\end{itemize}

Our action space spans multiple categories of operations:  
\begin{itemize}
    \item \textbf{GUI Actions}: direct interface manipulation following UI-TARS~\citep{qin2025ui}, e.g., clicks for element selection, typing for text input, and scrolling for navigation. Gameplay interactions also reuse these same primitives.  
    \item \textbf{Pre-defined SDK Functions}: supplementary operations that extend beyond GUI manipulation, including direct terminal commands for file management and software development, as well as MCP tool invocations for orchestrating external services and multi-tool reasoning. 
\end{itemize}

We define a \textit{step} as one complete ReAct cycle $(t_t, a_t, o_t)$. A trajectory of length $T$ is then formulated as:  
\begin{equation}
\tau = \{(t_0, a_0, o_0), (t_1, a_1, o_1), \ldots, (t_T, a_T, o_T)\}.
\label{eq:trajectory_definition}
\end{equation}

A key component of this formulation is the hierarchical memory state:  
\begin{equation}
\mathcal{M}_t = (\mathcal{W}_t, \mathcal{E}_t),
\end{equation}
where \textbf{Working Memory} $\mathcal{W}_t$ stores recent steps $(t_{t-k}, a_{t-k}, o_{t-k})$ in high fidelity for short-term reasoning, while \textbf{Episodic Memory} $\mathcal{E}_t$ maintains semantically compressed summaries of past episodes, preserving key intentions, and outcomes.
To remain efficient under long trajectories, we restrict direct context to the last $N$ steps from $\mathcal{W}_t$, while conditioning on $\mathcal{E}_t$ for longer-term recall. At each timestep, the policy predicts the next thought and action as:  
\begin{equation}
P(t_n, a_n \mid \text{instruction}, \mathcal{W}_n, o_n, \mathcal{E}_n).
\end{equation}

This highlights that agent behavior arises not from isolated predictions, but from an evolving loop of reasoning, action, feedback, and memory integration.

\subsection{Environment: All-in-One GUI Sandbox}

Training a general-purpose computer agent that seamlessly integrates a a wide range of computational capabilities imposes exceptionally demanding environmental requirements. Unlike single-domain simulators, such new environments must support diverse task types, integrate heterogeneous tools, and preserve long-lived state across complex, multi-step interactions.

To address these challenges, we engineered a universal sandbox that merges GUI operations and SDK functions (e.g., file system and tool calling) into a cohesive and versatile platform. A core innovation is the shared file system, which allows an GUI agent to, for instance, download a file via the browser and immediately process it using shell commands within the same containerized instance. The sandbox maintains the stability and reproducibility essential for complex tasks and allows for not only high-throughput training on a distributed computing backbone, but also a consistent environment for annotation, evaluation, and inference. Here we highlight the design of GUI and game sandbox.

\paragraph{GUI Env: Cloud Virtual Machine}  
To support large-scale training and evaluation of GUI agents, we developed a distributed virtual machine (VM) platform that runs mainstream desktop operating systems (Windows and Ubuntu) as well as the Android mobile OS. 
The platform integrates PyAutoGUI and ADB interfaces, enabling cross-device operations with minimal adaptation overhead. 
A unified SDK standardizes the entire interaction pipeline—from VM allocation and initialization to agent interaction, observation collection (e.g., screenshots and recordings), and task evaluation—making the system suitable for diverse use cases such as manual data annotation, OSWorld benchmarking, and online reinforcement learning.  

At the infrastructure level, the VM cluster comprises several thousand instances, centrally managed by a VM Manager capable of sustaining throughput at several thousand QPS (Queries Per Second) and handling high-concurrency execution. 
Each session is tracked with a task–environment mapping via session IDs to ensure state consistency across multi-round interactions. 
For monitoring and control, all sessions are visualizable in real time via VNC (Virtual Network Computing) / RTC (Real-Time Communication). 
A lease-based lifecycle mechanism automatically releases resources after task completion or failure, while overdue sessions are reclaimed to prevent waste.  

Beyond GUI interaction, the platform extends the agent’s capabilities with tool calling and coding support, enabling cross-domain workflows such as web browsing, file manipulation, and software development. 
An integrated endpoint pre-loads essential local services for browsing, file access, and terminal use, ensuring that tools are available out-of-the-box. 
The sandbox also enhances the coding environment by allowing services launched from the terminal to be exposed via proxy URLs, enabling the GUI agent to preview both front-end and back-end components. 
For human-in-the-loop debugging and annotation, the environment further provides VNC, a remote VS Code editor, Jupyter, and terminal previews directly in the browser.

\begin{figure*}[t]
    \centering
    \includegraphics[width=0.98\linewidth]{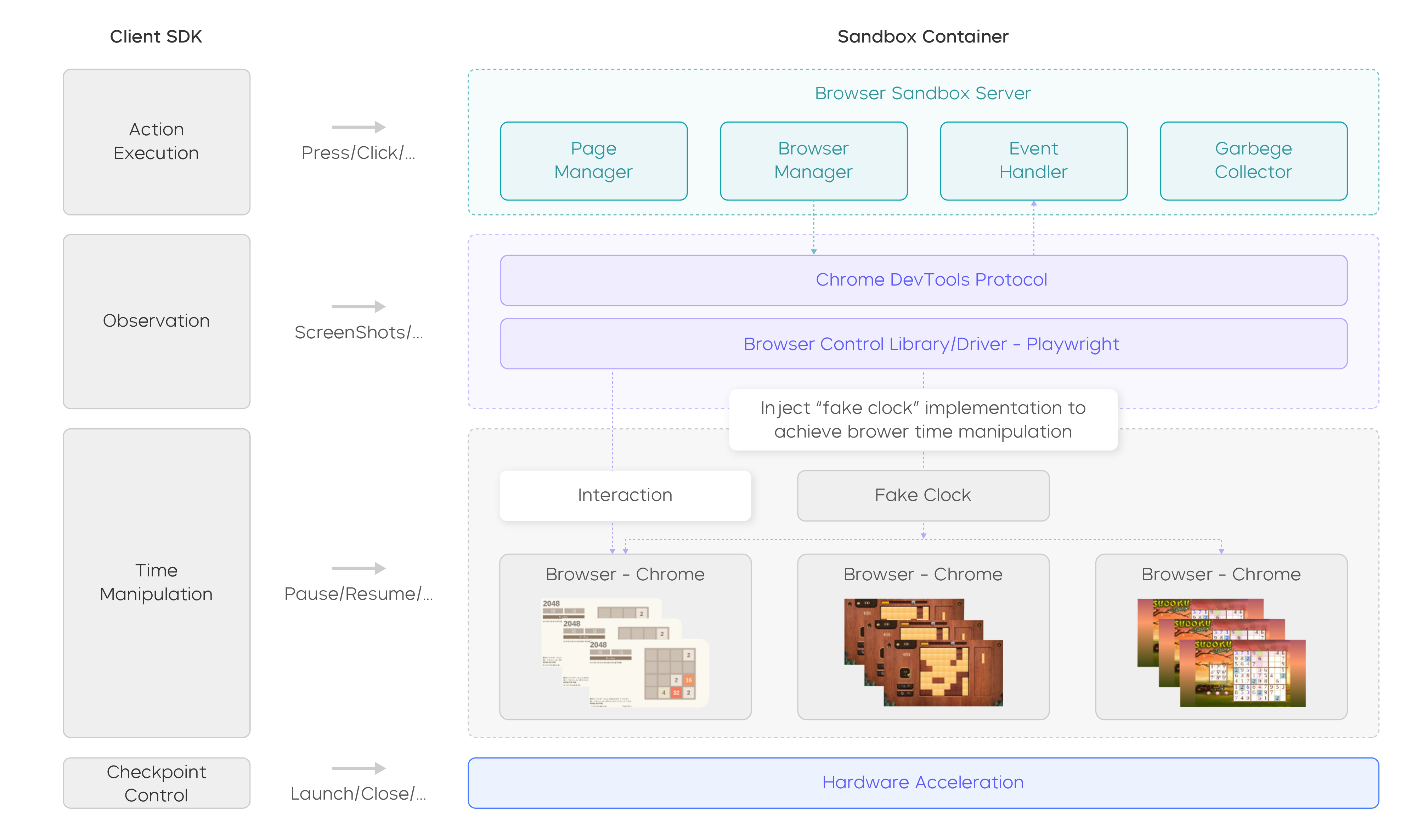}
    \caption{
    Browser sandbox (container) architecture.
    }
    \label{fig:sandbox}
\end{figure*}

\paragraph{Game Env: Hardware-Accelerated Browser Sandbox}  
To support high-throughput rollouts for multi-turn RL on web-based mini-games, we built a browser sandbox that serves as the execution and observation backbone. 
Because these mini-games run entirely in HTML5/WebGL, a browser environment is the only practical way to execute them faithfully while capturing their full interactive state. 
The sandbox exposes unified ``page management + page interaction'' APIs: clients issue actions (e.g., keyboard/mouse inputs) and receive synchronous observations (screenshots, scores, levels), completing the standard action-to-state loop.  

As illustrated in Figure~\ref{fig:sandbox}, concurrency is achieved by running multiple browser instances per container with elastic scheduling. 
The system monitors main processes and performs automatic crash recovery to ensure long-running stability. 
A page-control layer manages page creation and deletion, maintains session–page mappings, tracks page states, and executes commands, while checkpointing ensures reproducibility. 
An event handler continuously reports browser/page events to the manager, and a garbage collector reclaims idle sessions to prevent resource leakage.  

For programmatic access, the sandbox is compatible with the Chrome DevTools Protocol and popular drivers such as Playwright, enabling orchestrated, debuggable, and auditable interaction. 
GPU-based hardware acceleration reduces screenshot overhead, while re-implemented Window timing APIs allow time acceleration and pause at startup, improving sampling efficiency and reproducibility without altering game logic.  
In sum, the sandbox functions like a standard RL environment but is engineered specifically for the web stack, balancing high concurrency, determinism, and reproducibility.

\subsection{Data Flywheel Overview}
\label{sec:data_fly_whell}
\begin{figure*}[t]
    \centering
    \includegraphics[width=0.85\linewidth]{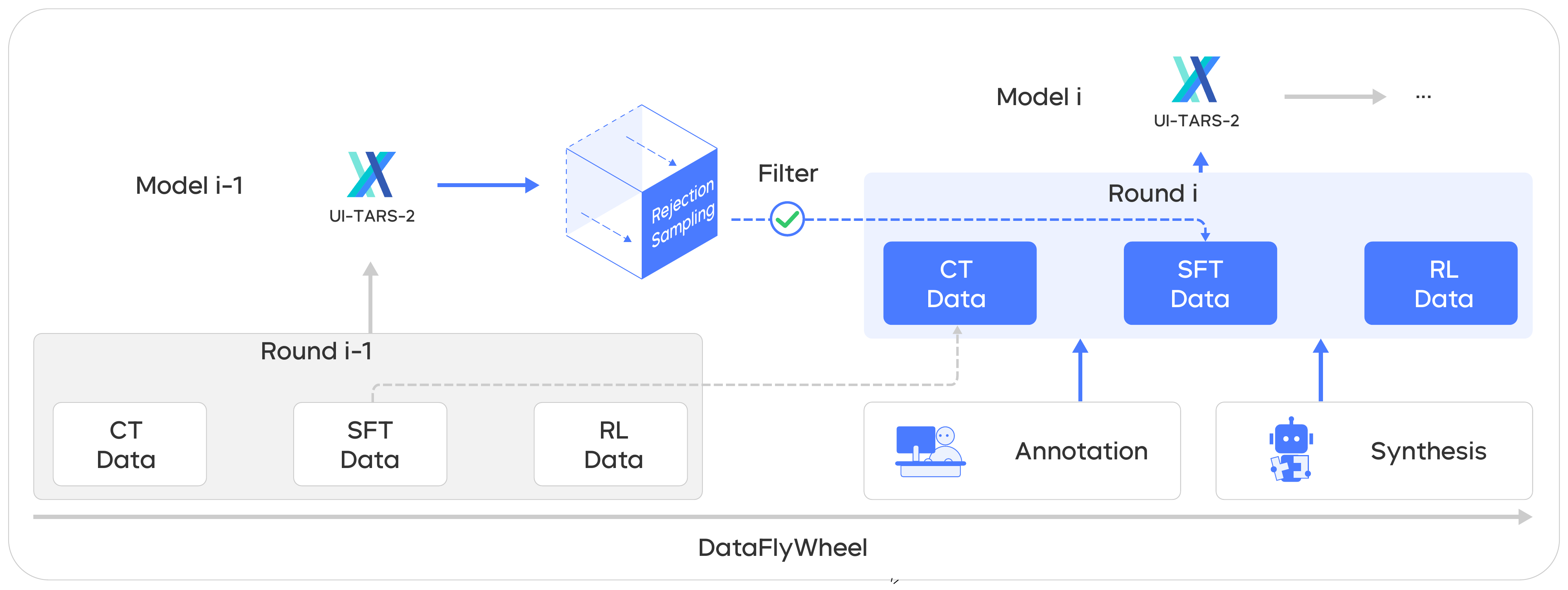}
    \caption{
    We curate a Data Flywheel for UI-TARS-2, establishing a self-reinforcing loop that continuously improves both data quality and model capabilities.
    }
    \label{fig:model}
\end{figure*}

As shown in Figure~\ref{fig:model}, we introduce the data flywheel that continually improves both model capabilities and data quality through repeated training cycles. In each cycle, the latest model generates new agent trajectories, which are filtered and redistributed to the most suitable training stages. High-quality outputs are promoted to later stages (e.g., SFT), while lower-quality outputs are recycled into earlier stages (e.g., CT). Over successive iterations, this dynamic reallocation ensures that every stage operates on optimally matched data, creating a self-reinforcing loop where better models yield better data, and better data produces better models.

\paragraph{Training Stages} 
Starting from the pre-trained checkpoints of Seed1.6~\citep{thinking1.6}, the flywheel operates through three stages: continual pre-training (CT) — broad knowledge acquisition from large-scale, diverse data, supervised fine-tuning (SFT) — high-quality, task-specific instruction tuning, and reinforcement learning — end-to-end optimization on verifiable interactive tasks. In each iteration, the current RL model generates new trajectories. High-quality outputs are appended to the SFT dataset, lower-quality ones are routed to CT, and the model is retrained sequentially on the updated CT, SFT, and RL stages.

\paragraph{Cold-start Data Sources} 
The flywheel is bootstrapped with two initial datasets. For CT, we collect task tutorials, instructional videos, demonstrations from the internet, and our in-house data (\cref{sec:in_situ_annotation}) to form the base knowledge set \( D_{CT}^{(0)} \). For SFT, we construct \( D_{SFT}^{(0)} \) through synthetic data generation and human annotation. During both CT and SFT, agent-specific data is mixed with general-purpose data, including chat and reasoning domains. Agent-specific data constitutes only a small fraction of CT, which emphasizes broad knowledge acquisition. In contrast, agent data forms a much larger proportion of SFT, which focuses on high-quality, task-specific agent trajectories.

\paragraph{Iterative Data Flow} 
After the initial RL model is trained, it becomes the main data generator for the next iteration. In each iteration \( t \), it produces new trajectories via rejection sampling (\textit{RFT}) or interactive annotation (\cref{sec:sft_data_construct}). Each sample is evaluated by a validation function \( V(s) \rightarrow \{0,1\} \). High-quality samples with \( V(s) = 1 \) are added to the SFT dataset for the next iteration as \( D_{SFT}^{(t+1)} = D_{SFT}^{(t)} \cup D_{\mathrm{RFT,high}}^{(t)} \), while lower-quality samples with \( V(s) = 0 \) are routed to the CT dataset as \( D_{CT}^{(t+1)} = D_{CT}^{(t)} \cup D_{\mathrm{RFT,low}}^{(t)} \). This ensures that SFT always receives the most recent, verified high-quality data, while CT continually expands with broader, less polished knowledge without contaminating the supervised signal. Note SFT and RL are performed more frequently than CT.
It should also be noted that in each cycle, we observe substantial transfer from general-purpose RL to agent-specific domains.
As iterations progress, the improved model \( M^{(t+1)} \) generates a higher proportion of high-quality outputs, i.e., \( P(V(s) = 1 \mid t) > P(V(s) = 1 \mid t-1) \), accelerating capability growth. Since every generated sample is reused at an appropriate stage, no data is wasted, creating a sustainable cycle in which model and data quality co-evolve to drive continual performance gains.

\subsection{CT \& SFT Data Preparation}

Agent-related training data represents a significant scarcity in existing human corpora, particularly for multi-turn interactive tasks that require sustained reasoning and tool manipulation. Unlike abundant mathematical or coding data in human corpora, agent interaction trajectories are rare and difficult to obtain at scale. To address this critical bottleneck, we develop a systematic data construction pipeline that operates on both interactive human annotations and automated data synthesis. 

\subsubsection{In-Situ Annotation for Continual Pre-training}
\label{sec:in_situ_annotation}
Our continual pre-training framework spans multiple agent domains. Here we illustrate the methodology using the GUI domain as a representative case. As the cold-start GUI CT dataset \( D_{CT, GUI}^{(0)} \), we include all training data from UI-TARS~\citep{qin2025ui} and UI-TARS-1.5~\citep{seed2025uitars15}, consisting of GUI tutorials collected from the internet, open-source agent trajectories, our in-house annotations, etc. Despite this diverse initialization, we quickly encountered several limitations. First, publicly available data is inherently scarce and easily exhausted, leaving insufficient coverage for training at scale. In particular, we observed a notable lack of content for Chinese-language applications, which hinders the development of truly versatile agents. Second, much of the available data provides only procedural actions while omitting the underlying cognitive reasoning. Models trained solely on such resources tend to mimic surface-level actions without internalizing the logic, leading to spurious or unstable reasoning chains. Ultimately, the central challenge of continual pre-training lies in how to systematically scale up high-quality, cognitively rich data to sustain long-term agent improvement.

To address the deficiencies of existing GUI datasets, we developed a large-scale, human-centric annotation system designed for collecting authentic cognitive processes. A key feature of our platform is its \textbf{in-situ deployment}: the annotation tool is directly installed on annotators' personal computers and runs unobtrusively alongside their normal usage. This design allows data to be collected continuously in realistic, everyday settings, without disrupting natural workflows.  

\textbf{Annotation Protocol.} An initial pilot study that attempted to retroactively add reasoning traces to recorded actions proved ineffective, as it was nearly impossible to reconstruct the annotator’s original thought process. Inspired by \citet{deitke2024molmo}, we instead adopted a \textit{think-aloud} protocol, where annotators verbalize their thoughts via audio while completing tasks. These verbalized thoughts are automatically aligned with corresponding UI interactions, producing data that captures both the reasoning chain and the grounded actions. To further enrich coverage, we recruited two groups of annotators: (1) \textbf{experts}, who provide demonstrations of complex tasks, and (2) \textbf{novices}, who are asked to solve unfamiliar tasks through exploration, trial-and-error, and external resources (e.g., web search). The novice track captures valuable data on problem-solving and adaptation when prior knowledge is absent.  

\textbf{Task Design and Collection.} To strengthen GUI-agent capabilities in realistic settings, we present a reproducible data acquisition pipeline. Candidate applications are selected using publicly available indicators along three dimensions—industry coverage, user engagement, and market penetration—yielding a representative set of mainstream websites and desktop applications. For each service, a hierarchical task graph is constructed, and task-importance scores are derived using normalized measures of usage frequency, user benefit, and cross-scenario transferability. We adopt a human–LLM collaborative workflow to generate multilevel query sets for each subfunction, spanning novice-to-expert skill levels and both single- and multi-application settings. A difficulty rubric based on step count, cross-page operations, prerequisites, and exception handling ensures balanced coverage across difficulty levels.  

\textbf{Curation Pipeline.} All collected data undergoes rigorous quality control, including executability verification, deduplication, and dual-annotator review. The audio-recorded thoughts are first transcribed using automatic speech recognition (ASR) and then refined by LLMs to produce coherent, high-quality reasoning text. These processed reasoning traces are precisely synchronized with on-screen actions, yielding temporally aligned reasoning–action trajectories. To further enhance training utility, we programmatically augment linguistic diversity and enrich reasoning chains, resulting in a final high-fidelity dataset suitable for continual pre-training.

\begin{figure*}[t]
    \centering
    \includegraphics[width=\linewidth]{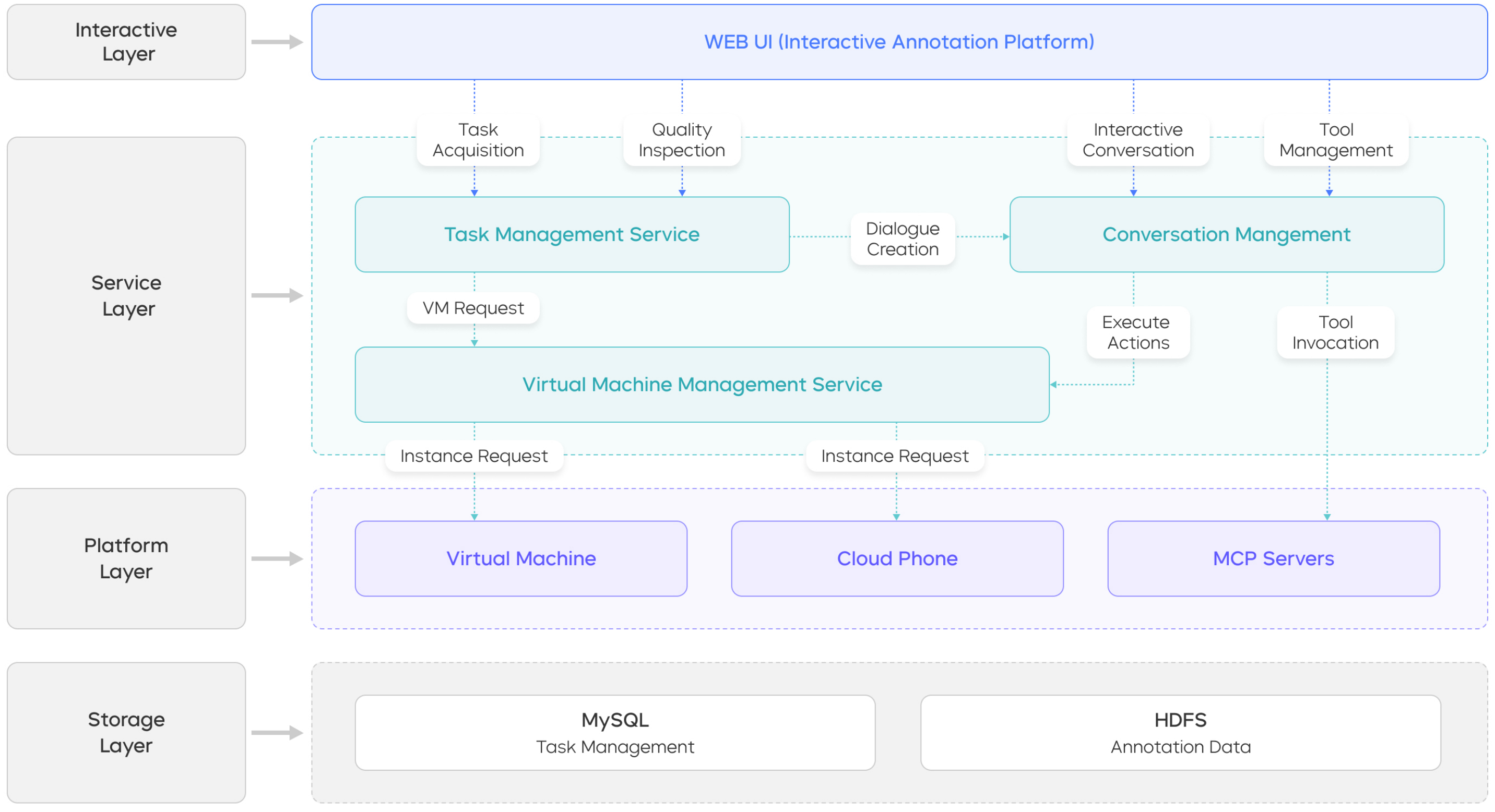}
    \caption{
    The four-layer architecture of the interactive annotation platform.
    }
    \label{fig:interactive_annotation}
\end{figure*}
\subsubsection{Interactive Annotation for Supervised Fine-tuning}
\label{sec:sft_data_construct}

A key challenge in training agents from human-generated SFT data is that such data is typically \textit{off-policy}: it does not reflect the actual distribution of actions that the model would take when interacting with an environment. As a result, models trained on this data may fail to generalize, since they never encounter or correct their own mistakes during rollout. Prior approaches mitigate this by asking annotators to correct errors in pre-collected trajectories~\cite{qin2025ui}. However, this procedure remains fundamentally offline and inefficient: it exposes model weaknesses only after task failure, without enabling real-time intervention or correction during interaction. Because agent training occurs within interactive environments, where actions directly affect subsequent states, this lack of on-policy supervision creates a significant gap. To bridge it, we propose a novel human-in-the-loop framework for online, interactive data annotation.

\paragraph{System Design}
Our interactive annotation platform is built on a four-layer architecture. At the top, the interaction layer presents the user interface, enabling annotators to engage with the system in real time. Beneath it, the service layer processes annotation requests, orchestrating model-generated command execution and human interventions. The platform layer provides scenario-specific execution environments—such as Computer Use, Phone Use, or Tool Use—tailored to different categories of tasks. Finally, the storage layer securely logs annotation data and complete interaction trajectories for downstream training and analysis. The overall design is illustrated in Figure \ref{fig:interactive_annotation}, which depicts the modular separation between layers and their control flow. In the following, we take GUI and Game as examples to illustrate the annotation process.

\begin{figure*}[t]
    \centering
    \includegraphics[width=0.9\linewidth]{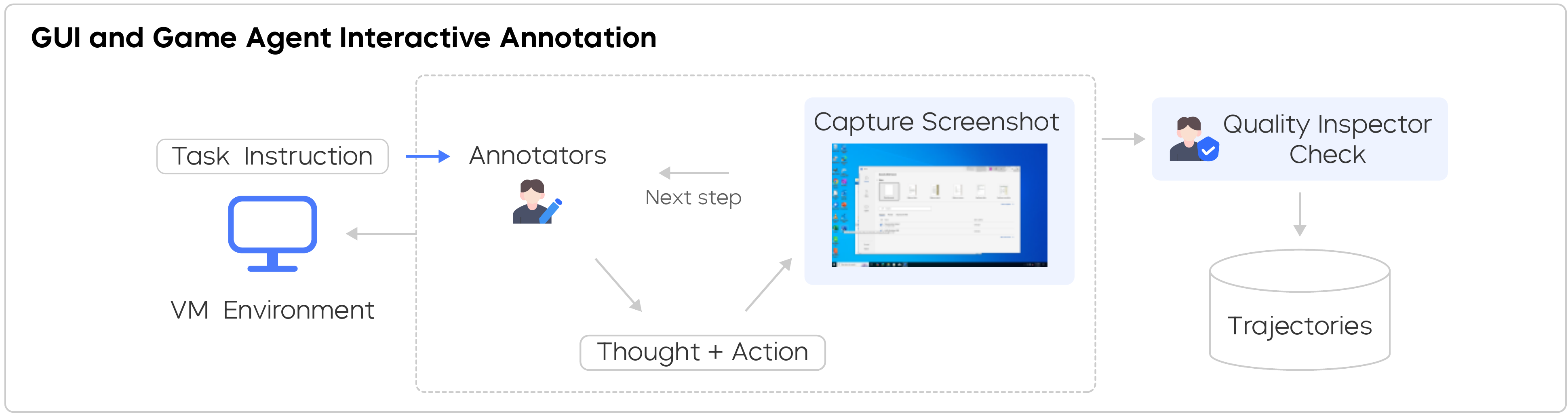}
    \caption{
    The interactive annotation workflow.
    }
    \label{fig:annotation_flow}
\end{figure*}

Our interactive annotation platform enables human annotators to provide online supervision directly within the agent's rollout. Annotators are assigned tasks to complete in a controlled virtual environment (see Figure~\ref{fig:annotation_flow}), backed by a cloud-hosted VM or browser sandbox to ensure reproducibility and consistent execution. At each decision point, the latest UI-TARS-2 model proposes candidate actions together with its reasoning trace. The annotator can either accept one of these suggestions or override it with a better thought and action, allowing human expertise to guide the trajectory in real time. We further streamline the workflow with features such as command auto-completion, real-time VM video streaming, and on-screen coordinate visualization, reducing latency and improving annotation accuracy.

Because annotation occurs in a live environment, annotators receive immediate feedback from the system and can track the evolving trajectory, avoiding the inefficiencies of post-hoc correction. This design ensures that all supervision remains strictly \textit{on-policy}: the data reflects the actual distribution of states visited by the current model. To further enhance efficiency, both the annotation model and the pool of tasks are periodically refreshed, ensuring that data collection consistently targets the weaknesses of the most recent agent.

\subsection{Multi-turn Reinforcement Learning}  
To train agents capable of long-horizon reasoning and interactive decision-making, we adopt a multi-turn RL framework built on RLVR (Reinforcement Learning with Verifiable Rewards)~\citep{guo2025deepseek}. 
We construct domain-specific pipelines that automatically synthesize large-scale, verifiable tasks across multiple domains. 
During RL, our model engages in real-time multi-turn interactions with the environment, continuously observing state transitions and environmental feedback until task completion. 
The model then leverages verifiable rewards to optimize its decision-making trajectories through iterative policy improvement. 
While our RL framework is applied across multiple domains with different tools defined by GUI operations and GUI-SDK functions, in the following, \textbf{we choose three representative cases to describe our framework}: (1) GUI-Browsing, which targets GUI-based information-seeking tasks, (2) GUI-General, which covers broader web manipulation tasks, and (3) gameplay, which focuses on lightweight web-based mini-games executed in a browser sandbox.

\subsubsection{Task Design} 
\label{sec:rl_task_design}

High-quality, sufficiently challenging, and verifiable task data for end-to-end RL remains extremely scarce. In the following, we introduce how to design training tasks that are both diverse in the form and equipped with reliable verification signals.

\paragraph{GUI-Browsing}  
To enable autonomous exploration in complex reasoning scenarios, we design an automated pipeline for synthesizing large-scale, verifiable GUI-browsing tasks. These tasks are conceptually similar to deep research tasks~\citep{openai-deepresearch}, except that agents must satisfy the information-seeking requirements solely through analyzing screenshots, without access to search APIs.
Our synthesis framework includes two main approaches: 

(1) Multi-Condition Obfuscation:
We begin by extracting core entities and their attribute features from authoritative knowledge sources (e.g., Wikipedia).  
Each feature is scored for distinctiveness using an LLM. Highly revealing attributes are removed, while the remaining ones are rewritten by the LLM to increase abstraction and reduce specificity.  
This process produces complex questions defined by multiple indirect constraints, requiring the model to combine and reason over blurred signals in order to identify the correct answer.  
For example, from a Wikipedia page we generate the following obfuscated question:  
``Discovered by a representative from the Music And Cabaret talent agency, this group had a founding lineup—initially under another name—that included members from Dreghorn and Irvine, plus a lead guitarist and drummer. The lead vocalist joined after being recommended by a founding member who saw them perform with a Kilmaurs-based band, and their lead guitarist left to form another ensemble before late 1975. Which record label did this group sign with?''  

(2) Multi-Hop Chain-Like Conditions:
We begin from an entity’s webpage and follow its hyperlinks to identify structurally related entities.  
For each linked entity, we extract and obfuscate descriptive features, creating tasks where the linked entity becomes the answer.  
We then treat the linked entity's page as the new starting point and repeat this process recursively, generating tasks for progressively deeper levels.  
At each step, the answer from the previous hop is embedded within the new question, forming a coherent reasoning chain.  
Finally, the atomic steps are semantically integrated into a single multi-hop question that requires the model to synthesize intermediate answers, mirroring the layered nature of knowledge propagation online and substantially increasing the demand for deep, sequential reasoning.  
To ensure difficulty, we filter the synthesized data by discarding instances that can be trivially solved using prior knowledge or a single-turn search, keeping only truly challenging and verifiable tasks for training.

\paragraph{GUI-General}  
To evaluate general-purpose interaction capabilities, we construct a dataset of GUI-General tasks using an offline synthesis pipeline centered on general websites.  
We begin by curating candidate websites from public collections, filtering out inaccessible pages, login-gated services, and trivial categories such as static information pages or casual games.  
For each selected website, VLMs are employed to identify and extract its core functionalities.  
Based on these, we synthesize tasks at the single-page level through a structured process: removing overly simple functions, composing executable instructions, merging prerequisite sub-tasks, and refining task descriptions for clarity, objectivity, and verifiability. 
The resulting dataset provides a diverse pool of executable, GUI-interaction-focused tasks that serve as queries for RL training, covering 690 websites across a wide range of domains.

\paragraph{Gameplay}
For the game domain, we construct the RL dataset through two complementary sources. First, we collect publicly available HTML5/WebGL mini-games that can run directly in the browser sandbox. Second, to further expand coverage, we synthesize new games using LLMs, which generate lightweight code implementations that preserve core gameplay mechanics while exposing explicit state interfaces. For both real and synthesized titles, we create concise JavaScript verification scripts that query runtime variables (e.g., score, level index, remaining lives) and provide time-aligned state attributes. These observations establish a reliable mapping from agent actions to environment transitions and reward signals. Finally, all interaction records are consolidated into a unified JSON schema containing scalar rewards, termination flags, and metadata (e.g., game version and verification checksums).

\subsubsection{Reward Design}
\label{sec:rl_verifier_design}
A reliable reward system is essential for stable policy optimization, requiring feedback signals that are both consistent and trustworthy across heterogeneous environments. 
We categorize our reward design based on whether the correctness of an agent's output can be deterministically verified:

\paragraph{Deterministically verifiable tasks}  
In domains where automatic function-based verifiers are available (e.g., games), 
we directly compute binary correctness signals as rewards.  
For GUI-Browsing tasks, where answers can be matched against reference ground truth, we instead employ \textsc{LLM-as-Judge}~\citep{gu2024survey} to evaluate the agent’s prediction against the target answer.  

\paragraph{Non-verifiable tasks}  
In more open-ended settings, such as GUI-General tasks, neither formal verifiers nor reference answers exist.  
To address this, we employ UI-TARS-2 as a generative outcome reward model (ORM) that produces scalar rewards conditioned on the agent’s trajectory.  
The ORM takes as input the full text history together with the last five screenshots (to fit within the context window) and outputs a score indicating task success.  
To achieve this, we specifically enhance UI-TARS-2's capability of ORM through targeted data annotation and single-turn RL, ensuring that its reward predictions are accurate, consistent, and robust for downstream multi-turn RL.

\begin{figure*}[t]
    \centering
\includegraphics[width=\linewidth]{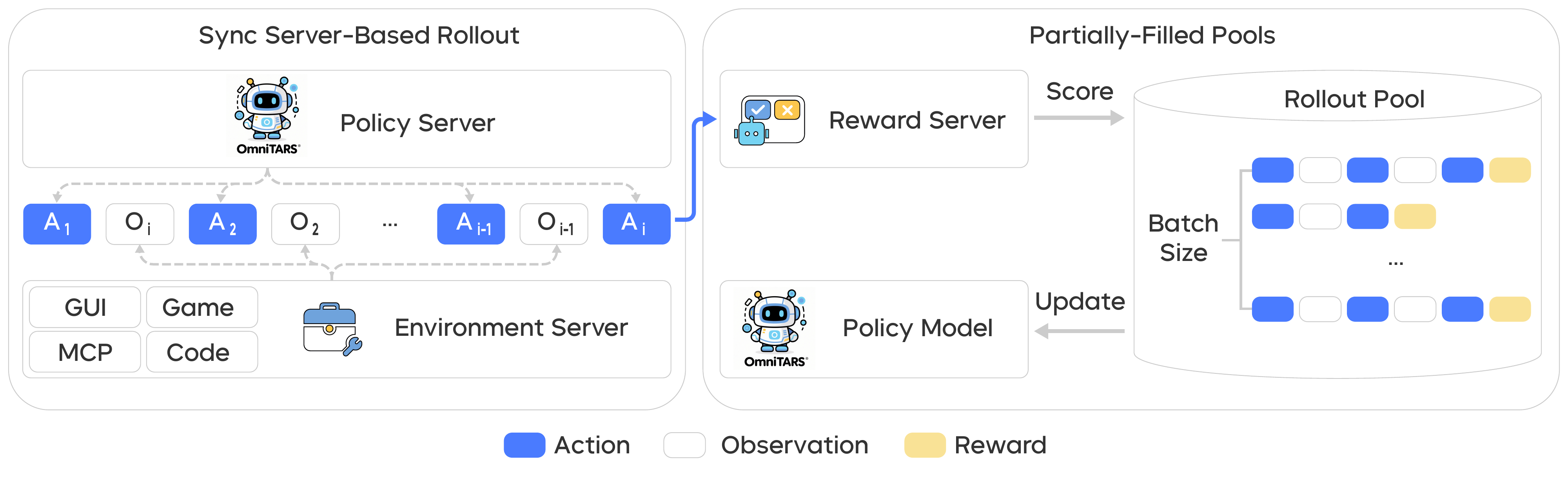}
    \caption{
    The multi-turn RL training infrastructure of UI-TARS-2.
    }
    \label{fig:multiturnrl}
\end{figure*}

\subsubsection{Asynchronous Agent Rollout via Stateful Environment}

Traditional batch-based rollout approaches often become bottlenecked by complex long-tail problems, reducing training efficiency and creating off-policy distribution drift. Our multi-turn RL training infrastructure (Figure~\ref{fig:multiturnrl}) is developed for two core objectives: (1) enhancing training stability and (2) optimizing efficiency in multi-turn rollout interactions and training sample organization. UI-TARS-2 implements several key features:

\paragraph{Asynchronous Inference with Server-Based Rollout}
We adopt a fully asynchronous inference system utilizing online server-mode processing. By encapsulating policy inference within asynchronous server architecture, we decouple agent reasoning framework implementation from policy inference execution. This design significantly enhances framework usability, which supports easily-developed new agent interaction handlers, while improving model inference efficiency through asynchronous inference. 

\paragraph{Streaming Training with Partially-Filled Rollout Pools}
Traditional batch-mode rollout requires complete batch inference before training initiation, potentially creating bottlenecks with long-tail cases that delay subsequent training cycles. Our system maintains a dynamic rollout pool where training updates commence once completed traces reach the minimum batch size threshold. Incomplete rollout traces remain in the pool for subsequent training iterations, ensuring continuous learning progress. This feature is conceptually similar to Kimi-Researcher~\citep{kimi-researcher}.

\paragraph{Stateful Agent Environment Integration}
We implement stateful agent environments that preserve execution states across multiple tool invocations, enabling continuous state transitions and maintaining context throughout extended problem-solving sessions. This approach supports complex, multi-step reasoning processes that require persistent environmental memory.

\subsubsection{RL Training Algorithm}
UI-TARS-2 is trained using Proximal Policy Optimization (PPO), where the policy is updated according to the following objective function:
\begin{equation}
\begin{aligned}
\mathcal{J}_\text{PPO}(\theta) = \mathbb{E}_{(q,a)\sim \mathcal{D},o_{\le t}\sim\pi_{\theta_{\text{old}}}(\cdot\mid q)}
\Bigg[ 
\min \Bigg( \frac{\pi_{\theta}(o_t\mid q,o_{<t})}{\pi_{\theta_{\text{old}}}(o_t\mid q,o_{<t})} \hat{A}_t, \\  
\ \text{clip} \Bigg( \frac{\pi_{\theta}(o_t\mid q,o_{<t})}{\pi_{\theta_{\text{old}}}(o_t\mid q,o_{<t})}, 1 - \varepsilon_{low}, 1 + \varepsilon_{high} \Bigg) \hat{A}_t \Bigg) \Bigg],
\label{eq:rl_alg}
\end{aligned}
\end{equation}
where $\pi_{\theta}$ is policy model, $\pi_{\theta_{\text{old}}}$ is previous policy model.

Following VAPO~\citep{yue2025vapoefficientreliablereinforcement} and VC-PPO~\citep{yuan2025whatspposcollapselongcot}, UI-TARS-2 integrates several critical enhancements to broaden the exploration space and improve stability, especially in long-horizon settings:

\paragraph{Reward Shaping}
To promote more strategic agent behaviors, the reward signal is mainly determined based on the correctness of the final outcome. In certain scenarios, we employ format rewards and length penalties to discourage premature termination or infinite continuation.

\paragraph{Decoupled GAE}
To address the challenge of value estimation bias over long sequences, we employ the Decoupled Generalized Advantage Estimation (Decoupled-GAE)~\citep{yuan2025whatspposcollapselongcot}, allowing the computation of advantage for the policy and value function to use different coefficients. Specifically, we set $\lambda_{policy}$ and $\lambda_{critic}$ to be different. This approach prevents decay in the critic's value estimates when dealing with lengthy token sequences, thereby promoting stability during long-horizon training.

\paragraph{Length-Adaptive GAE}
To mitigate the issue of inconsistent advantage estimation for sequences of varying lengths, we employ Length-Adaptive Generalized Advantage Estimation (Length-Adaptive GAE)~\citep{yue2025vapoefficientreliablereinforcement} technique, adjusting the GAE parameter $\lambda_{policy}$ based on the sequence length. Specifically, we set $\alpha=0.05$ in length-adaptive formula $\lambda_{policy}=1-\frac{1}{\alpha l}$ to control the overall bias-variance trade-off.

\paragraph{Value Pretraining}
To mitigate the value initialization bias, we adopt the Value-Pretraining~\citep{yue2025vapoefficientreliablereinforcement}, which involves offline training of the value model to convergence under a fixed policy. Specifically, responses are sampled continuously from a fixed policy (e.g., $\pi_{sft}$), and the value model is updated using GAE with $\lambda = 1.0$ (equivalent to Monte Carlo return), providing stable and reliable optimization. Training continues until crucial metrics such as value loss and explained variance reach sufficiently low levels, indicating effective convergence. The resulting value model checkpoint is then used as the initialization for subsequent experiments, ensuring more accurate and calibrated value estimation from the outset.

\paragraph{Clip Higher}
To further promote exploration, we decouple the PPO clipping parameters as recommended by DAPO~\citep{yu2025dapoopensourcellmreinforcement}, introducing distinct lower ($\varepsilon_{low}$) and upper ($\varepsilon_{high}$) clipping bounds. Increasing $\varepsilon_{high}$ affords greater flexibility for raising the likelihood of low-probability actions, thus enlarging the exploration space. Conversely, $\varepsilon_{low}$ is maintained at a low value to avoid prematurely eliminating tokens, which would risk collapsing the diversity of potential outputs.

\subsection{Merging Vertical Agents via Parameter Interpolation}
A central goal of UI-TARS-2 is to develop a unified digital agent that not only handles structured desktop and web interfaces but also extends to dynamic environments.  
A natural approach would be to conduct joint RL across all environments and tasks. However, this is challenging in practice: domains differ substantially in action/state spaces, task horizons, and rollout complexity, making large-scale joint optimization unstable and computationally prohibitive.  
Instead, we adopt a simpler but effective strategy that leverages the observation that models fine-tuned from the same pre-trained checkpoint remain approximately linearly mode-connected in parameter space~\citep{qin-etal-2022-exploring}.  
This property enables us to train specialized agents independently for different domains and then merge them through parameter interpolation, thereby consolidating their strengths without the cost of multi-domain joint training.  

Concretely, starting from a shared SFT initialization, we conduct multiple RL runs tailored to different environments—for example, \textbf{GUI-Browsing} tasks focused on information seeking, \textbf{GUI-General} tasks covering broader web manipulation, and \textbf{Game} environments based on interactive mini-games—alongside additional variants trained on other domains and corresponding tools (e.g., GUI-SDK).  
We then merge these trained models by interpolating their parameters:
\begin{equation}
\theta^{(\mathrm{merge})} = \sum_{k \in \{\mathrm{GUI\text{-}Browsing}, \mathrm{GUI\text{-}General}, \mathrm{Game}, \mathrm{GUI\text{-}SDK}, \ldots\}} \alpha_k \cdot \theta^{(k)}, 
\quad \text{s.t.} \quad \sum_{k} \alpha_k = 1, \ \alpha_k \ge 0,
\end{equation}
where $\theta^{(k)}$ denotes the parameters of each domain-specialized model.
Empirically, this interpolation strategy preserves the performance of each specialized vertical while enabling strong cross-domain generalization. On composite tasks requiring skills from multiple domains, the merged model performs almost comparably to the best specialized model in each relevant domain, without additional optimization cost.
\newcommand{\gemini}{Gemini 2.5 Pro}
\newcommand{\openaio}{OpenAI o1}
\newcommand{\claude}{Claude 3.7 Sonnet}
\newcommand{\openaigpt}{OpenAI GPT-4o}
\newcommand{\qwen}{Qwen 2.5-VL 72B}
\newcommand{\openaicua}{OpenAI CUA}

\section{Experiments}
\label{sec:experiments}
This chapter presents a comprehensive experimental analysis of UI-TARS-2.  
Although the training spans multiple domains and tool integrations, we focus our discussion on two representative settings: GUI-based interaction and game environments.  
These two cases highlight complementary challenges: structured interface operation on the one hand, and dynamic long-horizon control on the other.

\subsection{Experimental Setup}
UI-TARS-2 is initialized from the pre-trained checkpoint of Seed-thinking-1.6~\citep{thinking1.6}, and leverages all of its post-training data.  
The architecture includes a 532M-parameter vision encoder and a 
Mixture-of-Experts (MoE) LLM with 23B active parameters (230B total).  
Building on this base, we conduct multiple iterative training cycles consisting of SFT, RL, and RFT, progressively refining the model’s capabilities.

We conduct evaluations across a diverse set of benchmarks that comprehensively assess agent capabilities:

\paragraph{GUI Benchmarks} We evaluate our model across a diverse suite of benchmarks spanning three categories: computer use, mobile use, and browser use.  
For \textbf{computer use}, OSWorld~\cite{xie2024osworld} provides 369 tasks across Ubuntu, Windows, and macOS with detailed configurations and evaluation scripts, while WindowsAgentArena~\cite{bonatti2024windows} adapts this framework to over 150 Windows-specific tasks.  
To assess deeper system-level capabilities, we also include TerminalBench~\cite{tbench_2025}, which measures proficiency in command-line environments, and SWE-Bench~\citep{jimenez2023swebench}, which evaluates repository-level software engineering tasks.  
For \textbf{mobile use}, AndroidWorld~\cite{rawles2024androidworld} offers 116 tasks across 20 mobile applications within a live Android emulator, with dynamic task variations generated via randomized parameters.  
For \textbf{browser use}, Online-Mind2Web~\cite{xue2025illusionprogressassessingcurrent} contains 300 realistic tasks across 136 websites, while BrowseComp-en~\cite{wei2025browsecomp} and BrowseComp-zh~\cite{zhou2025browsecomp} provide high-difficulty multi-hop questions. 
For the above benchmarks, UI-TARS-2 is allowed to use either GUI operations or GUI SDK.

\paragraph{Game Benchmarks} 
We develop a \textbf{15 Games Collection} from our game pool, which is used to measure in-domain performance.  
We also leverage an OOD benchmark: \textbf{LMGame-Bench}~\citep{hu2025lmgamebenchgoodllmsplaying}, which evaluates LLM agents' game-playing abilities across six classic titles through a unified Gym-style interface, with optional perception and memory scaffolds designed to stabilize vision and long-horizon control.  
It reports performance under both harnessed and unharnessed settings.  
For all these collections, evaluations are conducted within a browser-sandboxed, screenshot-only setting.  
UI-TARS-2 interacts with games through a human-like action space (mouse clicks, key presses, and scrolling), mirroring how players operate in real environments.  
Results are reported as raw per-game scores as well as the mean normalized score across titles.  

\paragraph{Compared Baselines}  
For GUI benchmarks, we compare UI-TARS-2 against state-of-the-art proprietary models, including Claude 4~\cite{anthropic2025claude}, OpenAI-o3~\cite{openai2025o3}, and OpenAI CUA-o3~\cite{openai_2025_cua_blog}, as well as previous UI-TARS variants.  
For game benchmarks, we evaluate Claude (Computer Use)~\cite{anthropic_2024_developing_computer_use}, OpenAI CUA-o3, OpenAI-o3, Gemini-2.5 Pro~\cite{comanici2025gemini}, and Claude 3.7/4~\cite{anthropic2025claude}.

\subsection{Main Results}

\paragraph{GUI Main Results}  
As shown in Table~\ref{tab:gui_main}, UI-TARS-2 establishes superior performance across a wide range of GUI-agent benchmarks.  
Compared to previous versions of UI-TARS and other strong baselines such as OpenAI CUA-o3 and Claude 4, our model demonstrates consistent improvements across computer use, mobile use, and browser use settings.  
In particular, UI-TARS-2 surpasses UI-TARS-1.5 on all reported benchmarks, achieving 47.5\% on OSWorld, 50.6\% on WindowsAgentArena, 73.3\% on AndroidWorld, and 88.2\% on Online-Mind2Web, highlighting the benefits of iterative training and reinforcement learning.  
\textbf{Benefits from GUI-SDK}:  
With the integration of extended SDK functions, UI-TARS-2 is further equipped to handle system-level tasks beyond surface-level GUI interaction.  
In this setting, the model achieves 45.3\% accuracy on Terminal Bench, 68.7\% on SWE-Bench, 50.5\% on BrowseComp-zh, and 29.6\% on BrowseComp-en.  
For comparison, when restricted to GUI-only operation, the scores on BrowseComp-zh and BrowseComp-en are 32.1\% and 7.0\%, respectively.  
This clear performance gap demonstrates that GUI-SDK augmentation enables the model to perform more complex reasoning and tool-use behaviors, equipping UI-TARS-2 with the broader skills expected of general computer-use agents.  
\textbf{OOD Generalization}:  
Most of the tasks of GUI-Browsing and GUI-General are browser-focused tasks, after RL training, the resulting model exhibits strong OOD generalization.  
On Online-Mind2Web, RL improves accuracy from 83.7\% (the SFT baseline in the final iteration) to 88.2\%.  
More strikingly, the RL-trained model transfers effectively to domains that were not the primary focus of training: for example, OSWorld improves by nearly 10.5\% (from 43.0\% to 47.5\%) and AndroidWorld by over 8.7\% (from 64.6\% to 73.3\%).  
These results highlight the ability of task-specific RL to induce broadly transferable skills, enabling GUI agents to perform reliably in previously unseen environments.

\begin{table}[!t]
    \caption{Performance on computer use, mobile use, and browser use benchmarks.  
``-'' indicates unavailable; \ding{55} denotes lack of ability; and $\dagger$ indicates results obtained with an extended action space that includes GUI-SDK. Terminal Bench results are reported on 75 out of 80 tasks due to compatibility issues with our internal environment.
Abbreviations: WAA (WindowsAgentArena), BC-en (BrowseComp-en), BC-zh (BrowseComp-zh), TB (Terminal Bench), SB (SWE-Bench).
}
    
    \centering
    \resizebox{\textwidth}{!}{
    \begin{tabular}{l|c|c|c|c|c|c|c|c}
    \toprule
    \rowcolor{gray!33} & \multicolumn{4}{c|}{\emph{\textbf{Computer Use}}} & \multicolumn{1}{c|}{\emph{\textbf{Mobile Use}}} & 
    \multicolumn{3}{c}{\emph{\textbf{Browser Use}}} \\
    \midrule     
    \textbf{Model} & \textbf{OSWorld}& \textbf{WAA} & 
    \textbf{TB} &
    \textbf{SB} &
    \textbf{AndroidWorld} & \textbf{Online-Mind2web} & \textbf{BC-zh} & \textbf{BC-en}  \\
    
    \midrule
    Claude-4-Sonnet & 43.9 & - & 39.2 & 72.7 & - & - & 22.5 & 14.7 \\
    Claude-4-Opus & - & - & 43.2 & 72.5 & - & - & 37.4 & 18.8 \\
    OpenAI o3 & \ding{55} & \ding{55} & 30.2 & 69.1 & \ding{55} & \ding{55} & - & 49.7 \\
    OpenAI CUA-o3 & 42.9 & - & \ding{55} & \ding{55} & 52.5 & 71.0 & - & - \\
    \midrule
    UI-TARS & 24.6 & - & \ding{55} & \ding{55} & 44.6 & - & \ding{55} & \ding{55} \\
    UI-TARS-1.5 & 42.5 & 42.1& \ding{55} & \ding{55} & 64.2 & 75.8 & \ding{55} & \ding{55} \\
    \uitars-2 & 47.5 & 50.6& 45.3$^\dagger$ & 68.7$^\dagger$ & 73.3 &  88.2 & 32.1 (50.5$^\dagger$) & 7.0 (29.6$^\dagger$) \\    
    \bottomrule
    \end{tabular}}
    \label{tab:gui_main}
\end{table}

\paragraph{Game Main Results}  
As shown in Table~\ref{tab:game_results} and Table~\ref{tab:lmgame_bench}, UI-TARS-2 demonstrates strong performance across both in-domain and out-of-domain game benchmarks.  
On the 15-game in-house suite—where scores are normalized to Human (100)—UI-TARS-2 achieves a mean normalized score of 59.8, reaching nearly 60\% of human-level performance on average.  
This represents a substantial margin over production systems such as OpenAI CUA and Claude Computer Use, outperforming them by +35.0 and +38.2 points, respectively.  
Notably, the model is already close to human level on several titles, including 2048 (91.0), Infinity-Loop (92.7), Tiles-master (82.7), Snake-solver (76.5), and Merge-and-double (75.2), and even surpasses human performance on Shapes (108.9).  
On the out-of-domain LMGame-Bench, UI-TARS-2 remains competitive with frontier general-purpose models.  
For example, it achieves 117.1 on 2048 (within 9\% of o3 at 128.2), ranks near the top on Candy Crush (163.2, above o3's 106.0 and second only to Gemini-2.5 Pro at 177.3), and performs strongly on Super Mario Bros. (1783.2 vs.\ 1955.0 for o3).  
While performance is weaker on Tetris and Sokoban, reflecting challenges in very long-horizon planning, the overall results show that UI-TARS-2 can transfer effectively to unseen game mechanics and environments.

\begin{table}[!t]
\caption{15 Games Collection results. The last row reports the mean normalized score, computed by dividing each game score by the human score and averaging across games.}
    \centering
    \resizebox{\textwidth}{!}{%
    \begin{tabular}{l|c|c|c|c|c}
    \toprule
    \rowcolor{gray!33}\textbf{Game} & \textbf{Human} & \textbf{UI-TARS-2-SFT} & \textbf{UI-TARS-2-RL} & \textbf{OpenAI CUA} & \textbf{Claude Computer Use} \\
    \midrule
    2048 & 1024.31 & 968.00 & 932.40 & 911.21 & 800.00 \\
    Emoji-sort-master & 5.15 & 2.90 & 4.50 & 1.80 & 1.40 \\
    Energy & 10.08 & 2.40 & 3.30 & 0.82 & 1.04 \\
    Free-the-key & 5.54 & 0.00 & 0.70 & 0.00 & 0.00 \\
    Gem-11 & 186.85 & 84.90 & 63.90 & 47.00 & 55.00 \\
    Hex-frvr & 5276.85 & 1952.00 & 2389.00 & 615.59 & 523.07 \\
    Infinity-Loop & 6.58 & 1.60 & 6.10 & 3.30 & 1.90 \\
    Laser-maze-puzzle & 17.83 & 2.70 & 5.60 & 1.40 & 1.40 \\
    Maze:Path-of-Light & 7.17 & 1.10 & 2.00 & 0.35 & 0.82 \\
    Merge-and-double & 790.31 & 519.00 & 594.40 & 102.33 & 212.70 \\
    Shapes & 5.42 & 4.60 & 5.90 & 0.90 & 0.24 \\
    Snake-solver & 3.92 & 2.10 & 3.00 & 0.23 & 0.20 \\
    Tiles-master & 3.75 & 3.20 & 3.10 & 1.47 & 1.56 \\
    Wood-blocks-3d & 4646.00 & 1900.00 & 2908.00 & 1814.00 & 1632.00 \\
    Yarn-untangle & 19.75 & 4.30 & 7.00 & 5.05 & 1.56 \\
    \midrule
    Mean Normalized Score & 100.00 & 44.27 & 59.77 & 24.73 & 21.61 \\
    \bottomrule
    \end{tabular}}
    \label{tab:game_results}
\end{table}

\begin{table}[!t]
    \caption{LMGame benchmark results (mean$\ \pm\ $std over runs over three runs).}
    \centering
    \resizebox{\textwidth}{!}{%
    \begin{tabular}{l|c|c|c|c|c|c}
    \toprule
    \rowcolor{gray!33}\textbf{Model} & \textbf{Sokoban} & \textbf{Super Mario Bros} & \textbf{Tetris} & \textbf{2048} & \textbf{Candy Crush} & \textbf{Ace Attorney} \\
    \midrule
    claude-3.5-sonnet-20241022 & $0.0\pm0.0$ & $1540.1\pm21.7$ & $12.3\pm2.5$ & $57.8\pm16.4$ & $17.0\pm18.1$ & $1.0\pm0.0$ \\
    claude-3-7-sonnet-20250219 (thinking) & $0.0\pm0.0$ & $1430.0\pm162.2$ & $13.0\pm0.0$ & $114.2\pm7.2$ & $126.3\pm69.1$ & $3.0\pm0.0$ \\
    gemini-2.5-flash-preview-04-17 (thinking) & $0.0\pm0.0$ & $1540.7\pm262.4$ & $19.0\pm4.6$ & $107.4\pm3.4$ & $97.7\pm36.1$ & $1.0\pm0.0$ \\
    gemini-2.5-pro-preview-05-06 (thinking) & $1.0\pm0.0$ & $1025.3\pm443.2$ & $12.3\pm3.1$ & $120.5\pm3.9$ & $177.3\pm64.9$ & $8.0\pm0.0$ \\
    llama-4-maverick-17b-128e-instruct-fp8 & $0.0\pm0.0$ & $786.0\pm462.6$ & $11.7\pm1.2$ & $44.6\pm11.8$ & $32.3\pm41.4$ & $0.0\pm0.0$ \\
    gpt-4.1-2025-04-14 & $0.0\pm0.0$ & $1991.3\pm1018.5$ & $13.0\pm1.7$ & $94.5\pm17.0$ & $101.0\pm120.2$ & $0.0\pm0.0$ \\
    gpt-4o-2024-11-20 & $0.0\pm0.0$ & $1028.3\pm656.0$ & $14.7\pm2.1$ & $70.4\pm12.5$ & $59.0\pm54.6$ & $0.0\pm0.0$ \\
    o1-2024-12-17 & $0.0\pm0.0$ & $1434.0\pm0.0$ & $13.0\pm0.0$ & $128.1\pm20.8$ & $90.0\pm0.0$ & $3.0\pm0.0$ \\
    o3-2025-04-16 & $2.0\pm0.0$ & $1955.0\pm0.0$ & $31.0\pm0.0$ & $128.2\pm0.0$ & $106.0\pm0.0$ & $8.0\pm0.0$ \\
    o4-mini-2025-04-16 & $1.3\pm0.6$ & $1348.3\pm178.1$ & $15.0\pm3.6$ & $97.6\pm29.2$ & $110.7\pm49.7$ & $2.0\pm0.0$ \\
    \midrule
    UI-TARS-2 & 0.3$\pm$0.0 & 1783.2$\pm$63.7 & 16.0$\pm$1.3 & 117.1$\pm$1.9 & 163.2$\pm$31.3 & 7.0$\pm$0.0 \\
    \bottomrule
    \end{tabular}}
    \label{tab:lmgame_bench}
\end{table}

\subsection{Detailed Analyses}
In the following, we present the detailed analyses of UI-TARS-2's RL training.

\begin{figure*}[!t]
    \centering
    \includegraphics[width=0.95\linewidth]{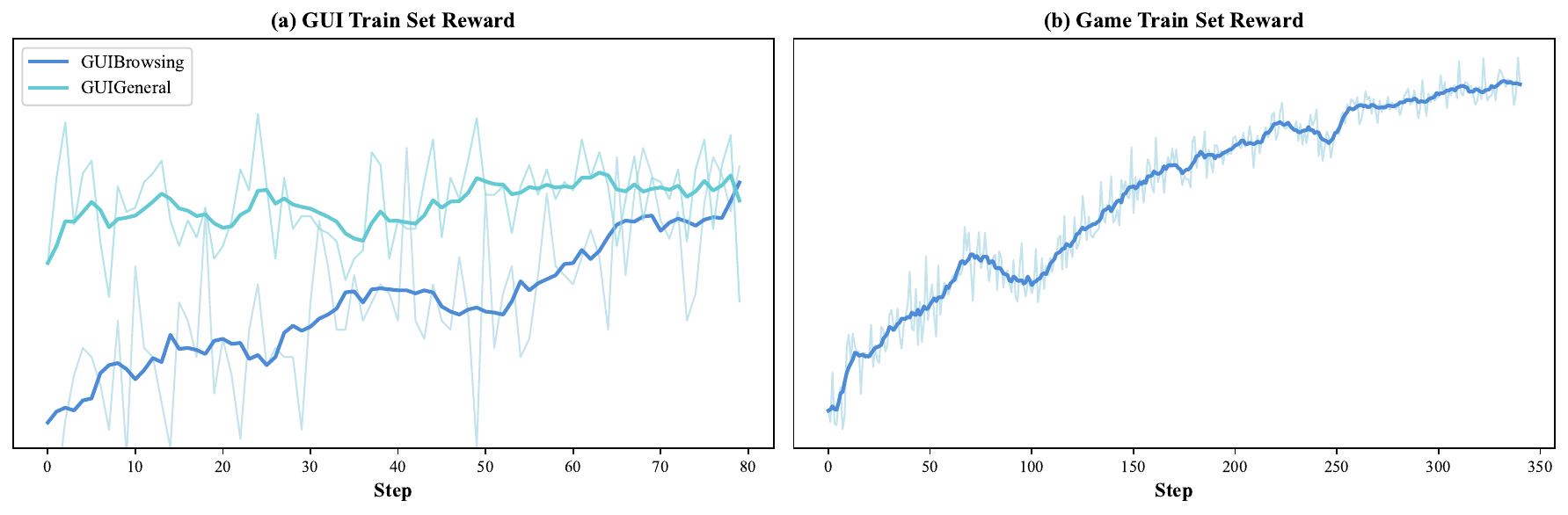}
    \caption{
   Training reward dynamics for GUI-Browsing, GUI-General, and game scenarios in UI-TARS-2.
    }
    \label{fig:reward}
\end{figure*}

\begin{figure*}[!t]
    \centering
    \includegraphics[width=\linewidth]{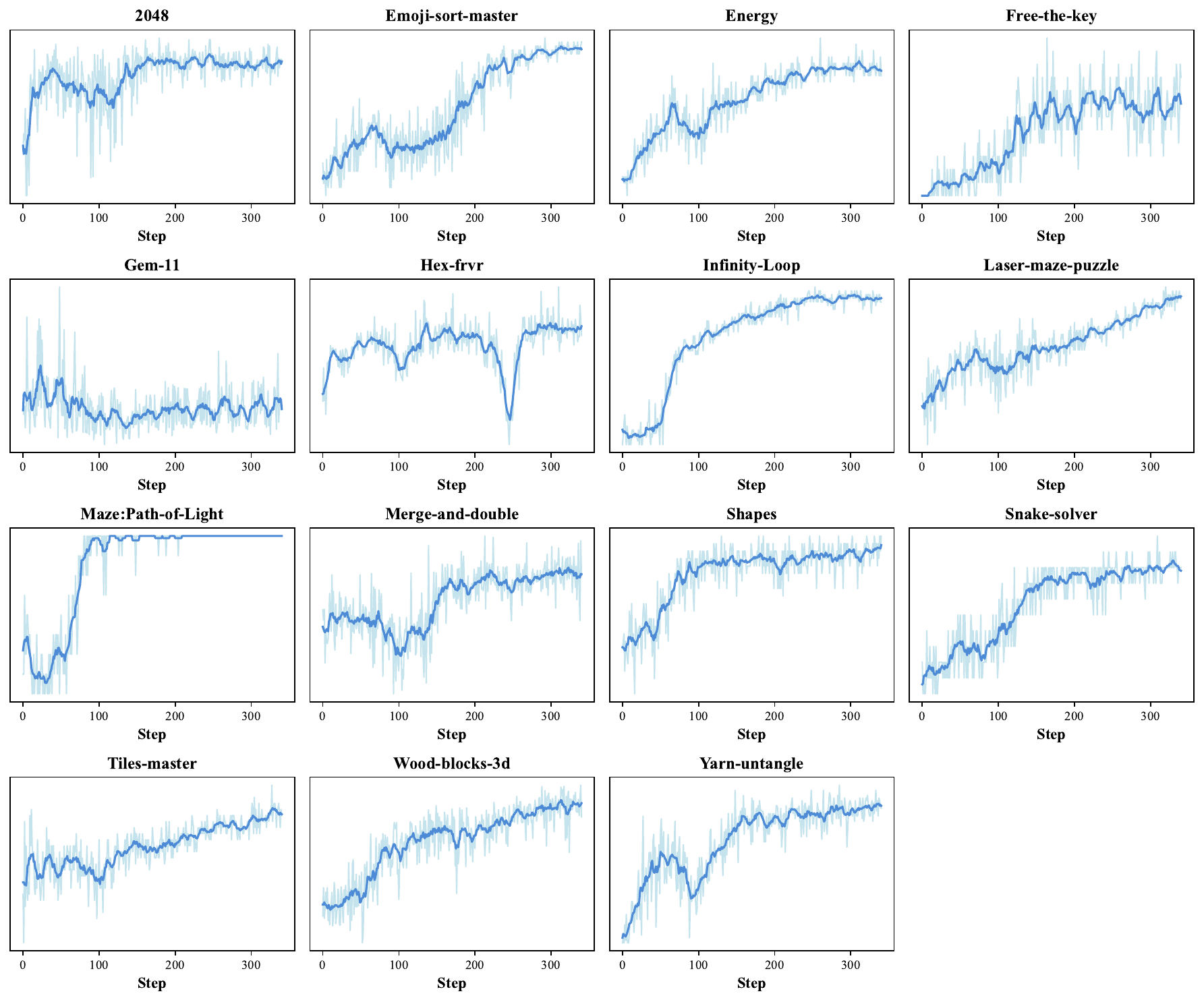}
    \caption{
    Training dynamics of training rewards for each game in the 15 Games collection.
    }
    \label{fig:per_game_reward}
\end{figure*}

\paragraph{Training Rewards and Entropy}  
As shown in Figure~\ref{fig:reward}, RL on GUI-Browsing, GUI-General, and Game tasks exhibits a clear upward trend in training rewards.  
For the game domain, we report the average reward over the 15-game suite, with per-game reward curves provided in Figure~\ref{fig:per_game_reward}.  
These results indicate that the policy steadily improves under RL supervision, and the consistency of this trend across both structured GUI tasks and dynamic game environments highlights the general effectiveness of multi-turn RL in diverse settings.  

\begin{figure*}[!t]
    \centering
    \includegraphics[width=0.95\linewidth]{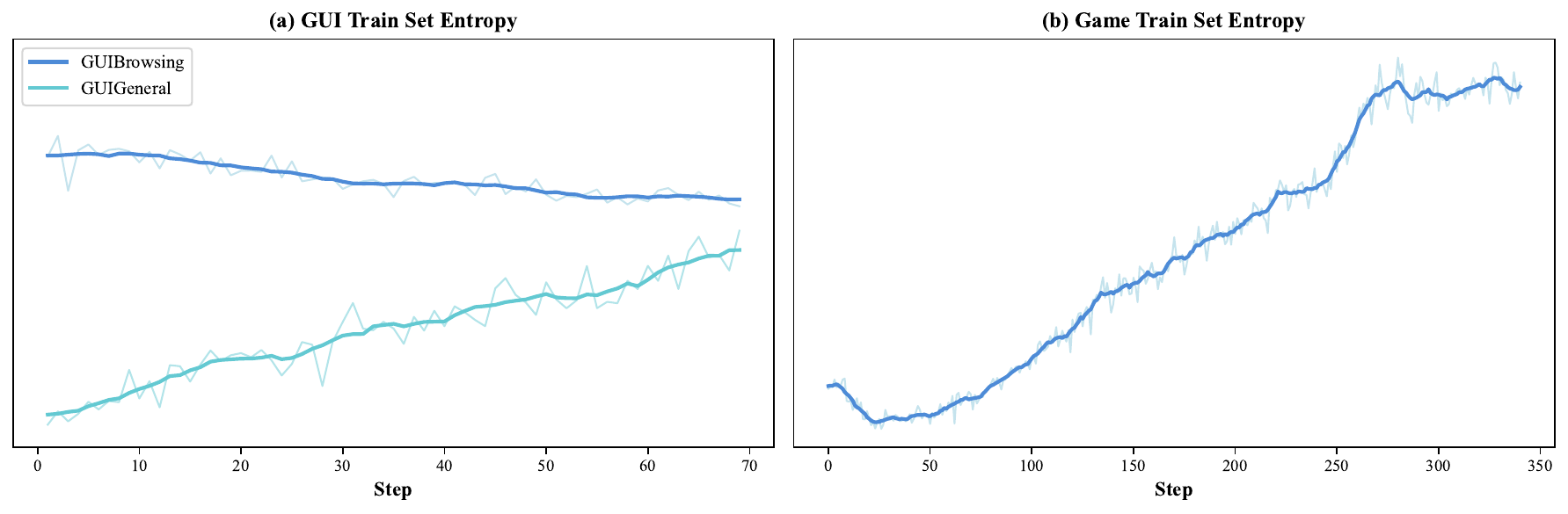}
    \caption{
  Training entropy dynamics for GUI-Browsing, GUI-General, and game scenarios in UI-TARS-2.
    }
    \label{fig:entorpy}
\end{figure*}

Interestingly, we observe distinct entropy dynamics compared to recent reasoning-focused RL work.  
Whereas reasoning RL often shows monotonic entropy reduction (reflecting a shift toward deterministic exploitation), our GUI and web-game experiments frequently exhibit rising entropy during training (Figure~\ref{fig:entorpy}).  
This suggests that the model maintains or even expands its exploration space as training progresses, enabling it to acquire new interaction patterns rather than collapsing prematurely into narrow exploitation.  
Such behavior reflects the need for broader exploration in visually rich and highly interactive environments, where diverse strategies are often required for effective learning.

\paragraph{Viability of VLM-as-Verifier} As shown in Figure~\ref{fig:reward}, UI-TARS achieves a steady increase in rewards on both training tasks, with the improvements correlating positively with performance gains on downstream benchmarks (Table~\ref{tab:gui_main}). Notably, although training was conducted with a generative reward model (for the web operation task) or GPT‑4o-as-judge (for the GUI browsing task), manual inspection of rewards did not reveal any substantial signs of reward hacking. This suggests that, unlike in general text generation tasks, employing a VLM-as-verifier for agent RL is feasible—possibly because task completion in agent settings can be defined more concretely and evaluated more objectively.

To more quantitatively examine the potential impact of reward hacking, we constructed an in-house ORM evaluation set containing 300 human-annotated GUI agent traces. On this benchmark, UI-TARS-2 achieved an F1 score of 83.8 as the generative ORM in the binary classification setting, indicating reasonably strong robustness. A closer analysis of misclassified cases reveals that the current ORM still exhibits a relatively high false positive rate. Nevertheless, even such an ``imperfect'' ORM proved effective in RL training. We attribute this to the fact that, even in a case where the final task outcome is incorrect, the agent might also execute many correct intermediate steps. In false positive cases, the model still receives appropriate rewards for these correct steps, and these positive contributions outweigh the erroneous rewards given to incorrect actions.

\begin{figure*}[!t]
    \centering
    \includegraphics[width=0.95\linewidth]{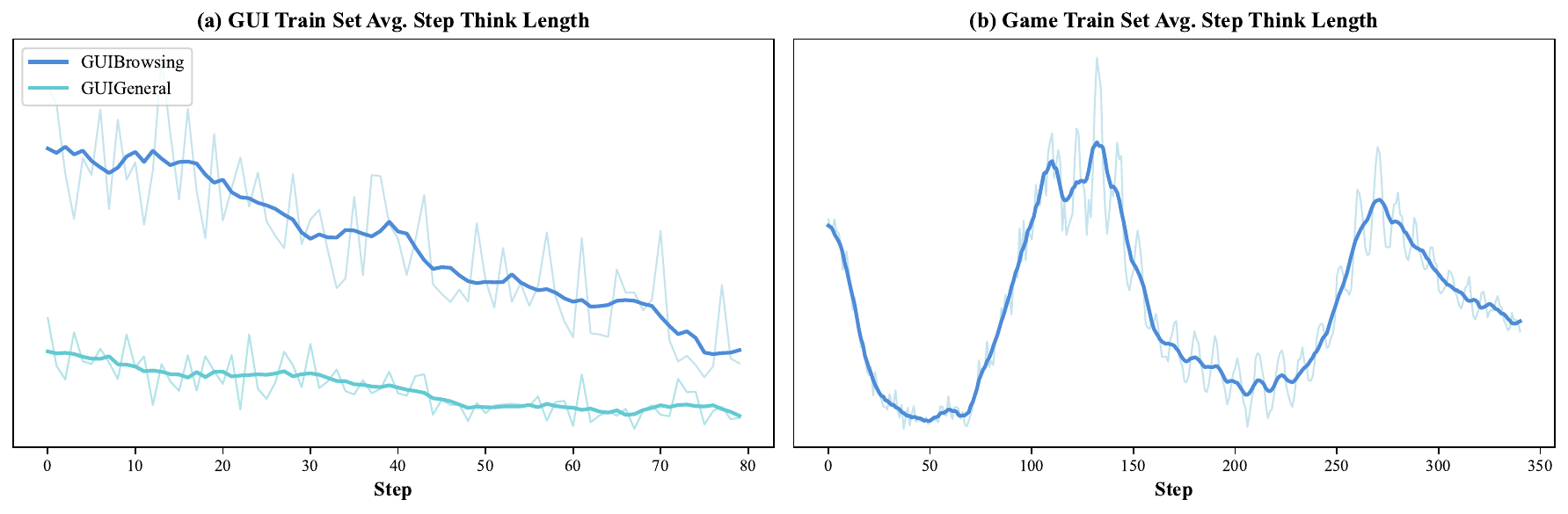}
    \caption{
    Training dynamics of average step think length for the GUI-Browsing, GUI-General, and game scenarios in UI-TARS-2 RL training.
    }
    \label{fig:token}
\end{figure*}

\paragraph{Average Think Length}
In our GUI experiments, we observe a consistent decline in the model's average step-level think length as RL training progresses (Figure~\ref{fig:token}). This trend stands in contrast to common expectations that more complex reasoning processes emerge over time. One possible explanation is that, in GUI tasks, agents primarily make progress through interaction with the environment rather than extended internal reasoning alone. Consequently, once the correct GUI action can be predicted, the agent can obtain rewards directly, reducing the need for longer deliberation. 

In the game domain, we observe a periodic pattern in the think length: it tends to increase for a period of time and then gradually decrease. Our analysis suggests that this pattern is tied to the progressive difficulty of the game (which is our design of the training curriculum, gradually increasing the game difficulty). Whenever the agent enters a new level, the increased difficulty requires more reasoning and decision-making to achieve success, leading to a rise in think length. As the agent becomes familiar with the challenges at a given difficulty level, the think length gradually decreases—similar to the trend observed in other GUI tasks—until the next difficulty escalation, at which point the cycle repeats.

\begin{figure*}[!t]
    \centering
    \includegraphics[width=0.95\linewidth]{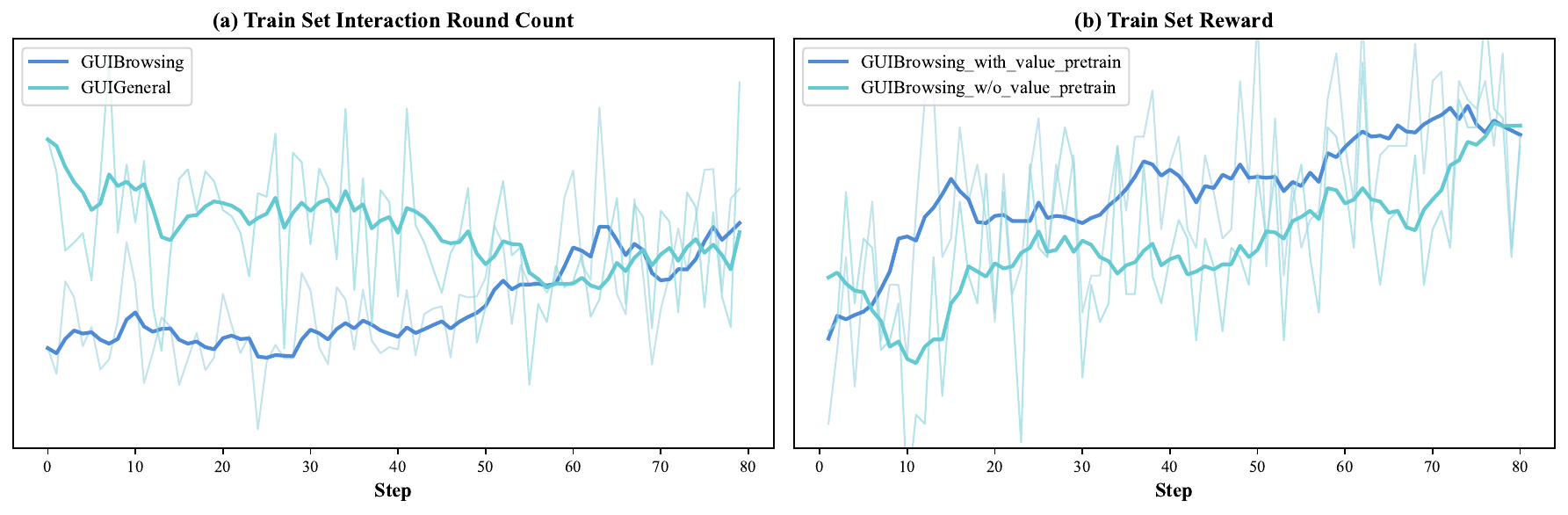}
    \caption{
     (a) Training dynamics of average interaction round count for the GUI-Browsing and GUI-General scenarios in UI-TARS-2 RL training;
     (b) Impact of value model pretraining in GUI-Browsing scenarios.
    }
    \label{fig:cnt}
\end{figure*}

\paragraph{Number of Environment Interaction Rounds}  
We observe that UI-TARS-2's interaction scaling curve—measured by the number of environment interaction rounds—is not always positively correlated with performance.  
As shown in Figure~\ref{fig:cnt}(a), while rewards steadily increase with training steps, the number of steps required to complete GUI-General tasks gradually decreases.  
This indicates that, through RL training, the model internalizes task-relevant knowledge and reduces unnecessary exploration, allowing it to solve tasks more efficiently.  
More broadly, interaction scaling is a common phenomenon in agent RL: models often learn to exploit larger budgets, prolonging trajectories before convergence.  
Such behavior can be mitigated by explicitly incorporating step budgets into the reward design, encouraging agents to balance efficiency with performance.  

\paragraph{Impact of Value Model Pretraining on PPO Training} In our preliminary experimental exploration, we observed that the value estimates of PPO-trained models were often negatively correlated with the obtained rewards. Motivated by this finding, we introduced a value model pretraining stage into the training process. 
As shown in Figure~\ref{fig:cnt} (b), value pretraining enhances the value model’s ability to guide policy learning, leading to consistently higher rewards throughout training.

\begin{figure}[!t]
    \centering
    \includegraphics[width=0.95\linewidth]{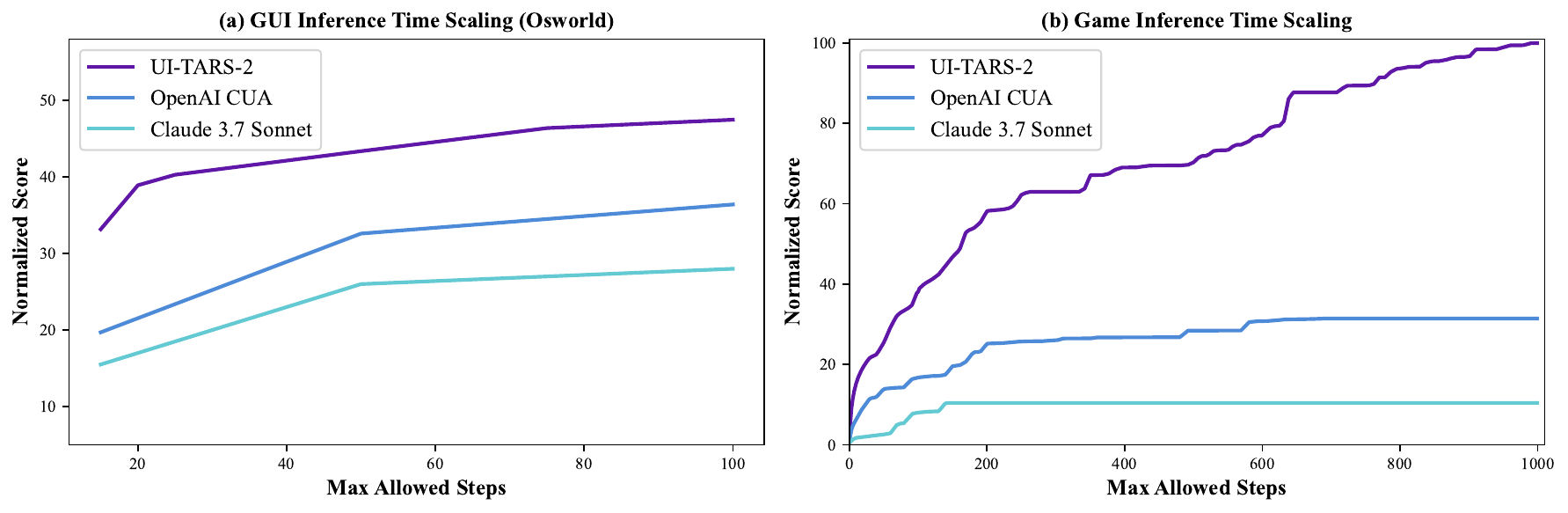}
    \caption{
    Inference-time scaling evaluation on OSWorld and game benchmarks.
    }
    \label{fig:inference_time_scaling}
\end{figure}

\paragraph{Inference-time Scaling}
The long-horizon nature of gameplay makes it a natural testbed for inference-time scaling. As the allowed step budget increases (Figure~\ref{fig:inference_time_scaling}, our performance curve rises steadily in an almost monotonic, staircase-like pattern, without exhibiting instability spikes. In contrast, baseline curves flatten quickly, indicating a limited ability to translate additional interaction budgets into further gains. The persistent upward trend of our agent, even at step counts orders of magnitude larger than the original budget, suggests that the policy continues to unlock new subgoals as task-specific thresholds are crossed, rather than merely looping or drifting.

We also observe a similarly strong inference-time scaling trend on both OS operation tasks (OSWorld), as shown in Figure~\ref{fig:inference_time_scaling}. The plots clearly indicate that as the maximum allowed inference steps increase, the model's performance score consistently rises, which serves as a key indicator of its capacity to leverage a larger computational budget for improved outcomes. Interestingly, although RL training has incentivized our agent to complete tasks in fewer steps, the model still exhibits excellent inference-time scaling on the OSWorld evaluation set. This finding highlights that the learned policy retains the flexibility to effectively exploit additional interaction steps at inference time—unlocking further subgoals or exploring alternative solution paths—rather than overfitting to minimal-step strategies during training.

\begin{figure*}[!t]
    \centering
    \includegraphics[width=0.95\linewidth]{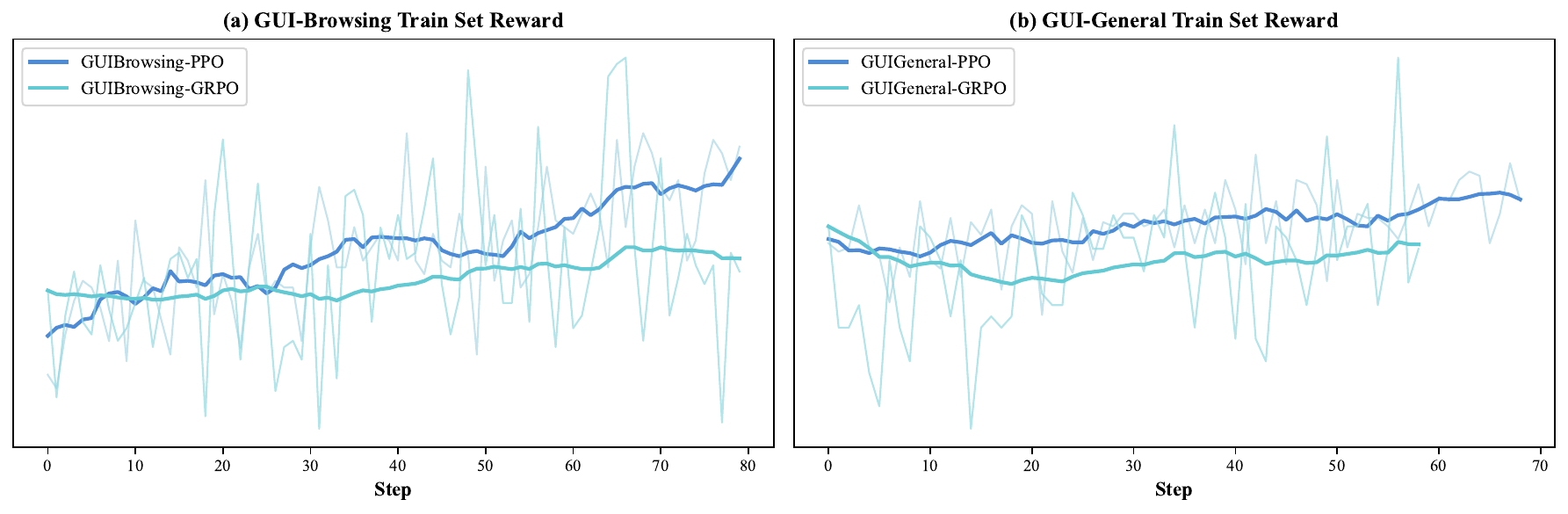}
    \caption{
   Training dynamics training reward of GUI-Browsing and GUI-General during PPO and GRPO training.
    }
    \label{fig:ppo_grpo}
\end{figure*}

\paragraph{PPO v.s. GRPO}   
GRPO~\citep{shao2024deepseekmathpushinglimitsmathematical} has been shown to be effective for training across a wide range of reasoning tasks. However, in our preliminary evaluation, we find that PPO consistently outperforms GRPO by a clear margin. As depicted in Figure~\ref{fig:ppo_grpo}, PPO maintains higher rewards with lower volatility throughout training. To ensure stable learning dynamics and achieve stronger overall performance, we ultimately selected PPO as the optimization algorithm for our main experiments.

\begin{figure*}[!t]
    \centering
    \includegraphics[width=0.95\linewidth]{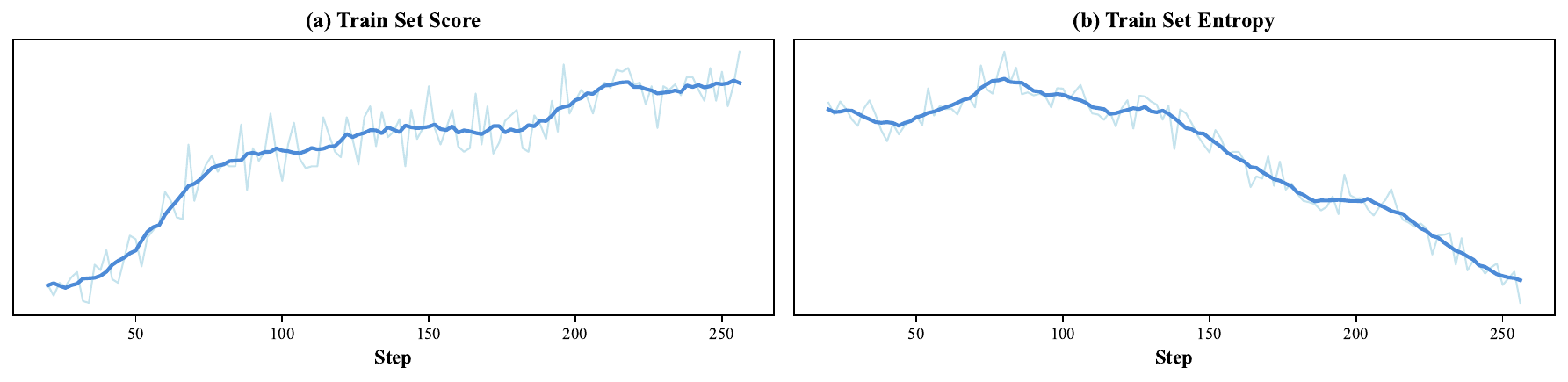}
    \caption{Training dynamics (train set score and entropy) of GUI-SDK RL.
    }
    \label{fig:mcp_exp_analysis_m8_main}
\end{figure*}

\paragraph{Behavior Analyses in Game RL}
In the Figure~\ref{fig:per_game_reward}, we present the training rewards for every game. Several titles reach or closely approach the human@100-step reference by the end of training (for example, 2048, Infinity-Loop, Emoji-sort-master, Tiles-master, and Shapes, with Shapes surpassing the human reference). A second pattern is from-zero learning: games that the base model could barely play at the start (such as Free-the-key and Yarn-untangle) are trained up to nontrivial scores, indicating a genuine increase in the model’s general game-reasoning ability rather than overfitting to a few scripts.

At the same time, a subset of games shows clear plateaus or temporary regressions followed by shallow recovery (for example, Gem-11 and Hex-frvr), suggesting a reasoning ceiling imposed by the starting backbone rather than a lack of optimization steps. The staircase shapes visible on many curves indicate that progress tends to arrive in bursts when task-specific subgoals become reliably attainable; once a subgoal is mastered, learning stabilizes until the next threshold is unlocked.

Overall, the curves imply that additional compute and curriculum can continue to convert training into control gains, but breaking through the remaining walls will likely require stronger long-horizon reasoning and planning capacity (for instance, better credit assignment, curriculum over subgoals, and improved search or memory components), pointing to clear headroom for future scaling.

\paragraph{Analysis of GUI SDK RL}
As shown in Figure~\ref{fig:mcp_exp_analysis_m8_main}, 
during GUI-SDK RL, 
the training score exhibits an overall increasing trend as the training steps progress. This indicates that the model gradually becomes proficient in utilizing external tools to solve the complex problems throughout the training process, seamlessly combining the reasoning and tool call in process.
Meanwhile, the training entropy shows a continuous downward trend as training progresses, suggesting that the model’s confidence in its predictions steadily improves, leading to enhanced stability and reduced uncertainty in its reasoning path. 

\begin{figure*}[!t]
    \centering
    \begin{subfigure}[b]{\textwidth}
        \centering
        \includegraphics[width=0.9\textwidth]{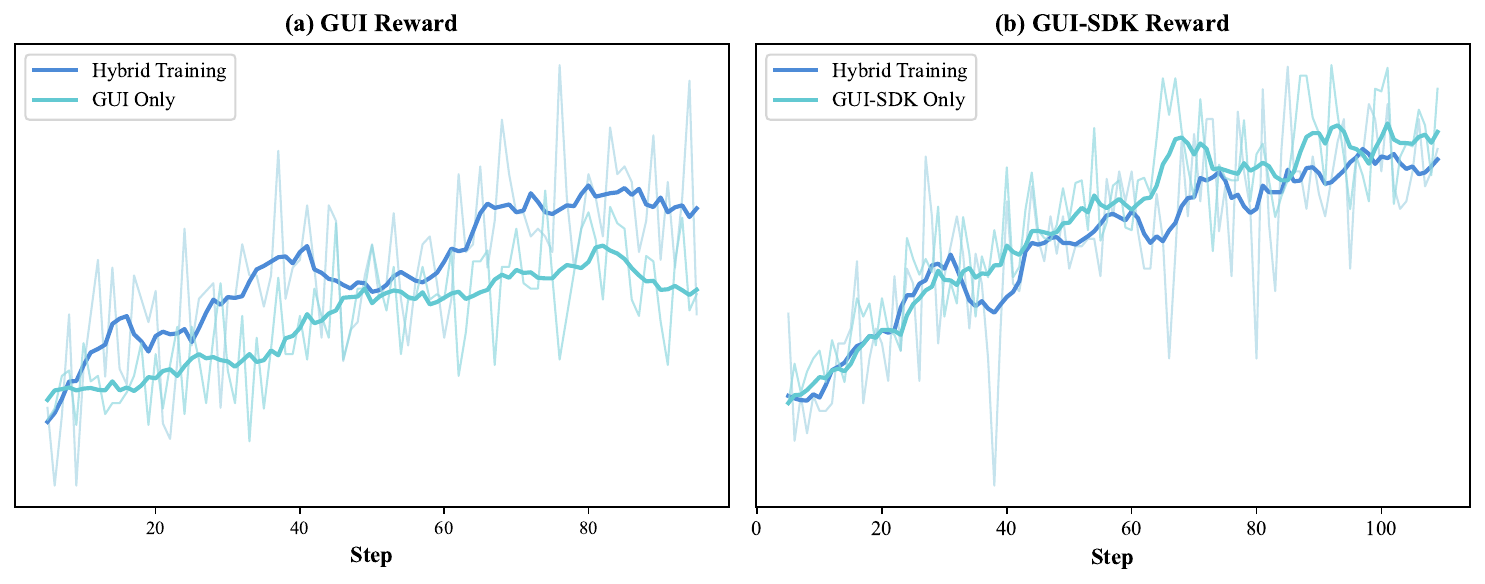} 

    \end{subfigure}
    
    \vspace{10pt} 
     \begin{subfigure}[b]{\textwidth}
        \centering
        \includegraphics[width=\textwidth]{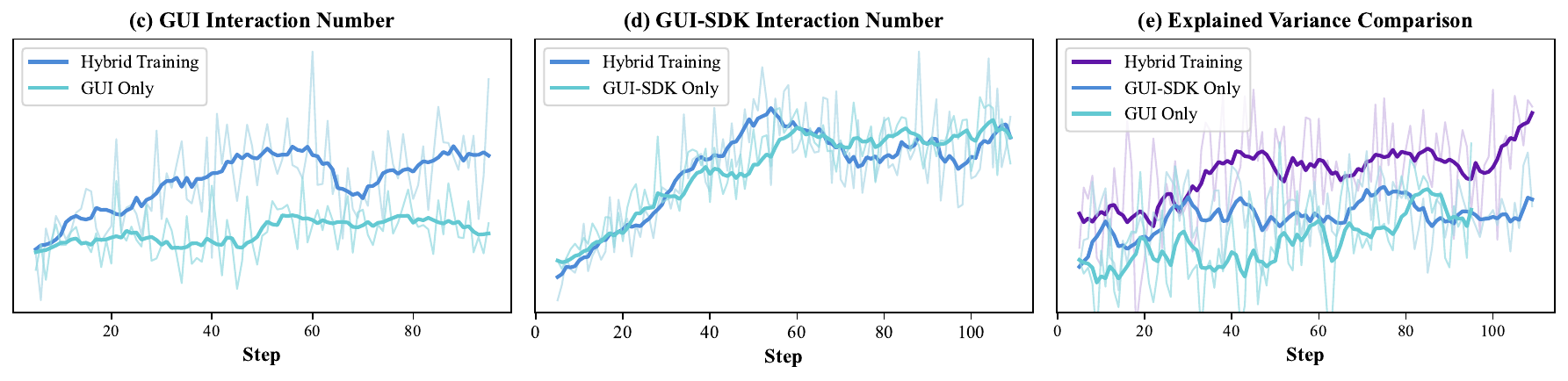} 
        \caption{Training reward of hybrid training versus training purely on one interface.}
    \end{subfigure}
    
    \caption{
    (a)(b) Training reward of hybrid training versus training purely on one interface;
    (c)(d) Iteration length of hybrid training (Orange line) with training purely on one interface (Blue line);
    (e) Explained Variance of the value model in hybrid training versus training on one modality. Unified tasks enable the value function to perform optimally across all scenarios.}
    \label{fig:hybrid-dynamic}
\end{figure*}

\paragraph{Hybrid Agent RL}
While parameter interpolation serves as our primary approach for consolidating specialized agents, we also investigated an alternative based on hybrid reinforcement learning.  
In this setting, we focused on an information-seeking scenario that could be solved through two distinct interfaces: a GUI-only method and a GUI-SDK method, where graphical actions are augmented with additional system-level capabilities.  
The hybrid model was trained to use either interface, whereas baseline models were trained exclusively on a single interface with the same overall batch size.  

As shown in Figures~\ref{fig:hybrid-dynamic}, hybrid training produced a stronger interaction scaling trend.  
Even though the training data for each interface was effectively halved compared to the single-interface baselines, the hybrid model outperformed the GUI-only baseline when evaluated on pure GUI tasks.  
This indicates that knowledge acquired through the more capable GUI-SDK interface transfers effectively to GUI-only interaction, boosting competence even in the restricted setting.  
We further observed that employing a shared value model improved training stability and reward estimation: by learning jointly from trajectories across both interfaces, the value model generalized to a broader range of patterns, yielding higher explained variance than interface-specific baselines (Figure~\ref{fig:hybrid-dynamic}(e)).  

Compared to parameter interpolation, which merges specialized agents without additional optimization, hybrid training enables more direct cross-interface knowledge transfer but incurs higher training cost.  
Together, these results highlight two complementary strategies for unifying capabilities across interaction settings: interpolation offers efficiency, while hybrid RL provides stronger transfer.

\paragraph{Quantization for Latency Reduction}  
We examine the effect of W4A8 quantization on end-to-end efficiency. 
W4A8 quantization reduces model weights to 4-bit precision and activations to 8-bit precision, enabling faster inference with only a modest loss in accuracy. 
For \uitars-2, quantization increases the token generation rate from 29.6 to 47 tokens/s and reduces the average end-to-end latency per interaction round from 4.0 to 2.5 seconds. 
On OSWorld, accuracy decreases slightly from 47.5 to 44.4, indicating that the efficiency–performance trade-off remains favorable. 
These results demonstrate that W4A8 quantization is a practical strategy for deploying GUI agents in latency-sensitive settings while preserving competitive task performance.

\section{Related Work}
\label{sec:related_work}
Early LLM-based agents were primarily generic systems driven by prompting recipes that couple reasoning and acting, or by tool-augmented interfaces. Representative examples include ReAct, which interleaves chain-of-thought with environment actions~\citep{yao2023react}, MRKL, a modular neuro-symbolic tool hub~\citep{karpas2022mrkl}, and Toolformer, which self-supervises API calls to external tools~\citep{schick2023toolformer}. In parallel, DeepMind's Gato demonstrated a single multi-task policy acting across diverse embodiments~\citep{reed2022gato}.  
Building on these ideas, research soon specialized into vertical domains with dedicated benchmarks and interaction environments. For web/GUI interaction, Mind2Web and WebArena provide realistic websites and tasks for web agents~\citep{deng2023mind2web, zhou2023webarena}, while OSWorld enables execution-based evaluation on desktop applications~\citep{xie2024osworld}. For software engineering, SWE-bench frames end-to-end repository-level bug fixing and has spawned a series of agentic systems such as SWE-agent with agent–computer interfaces (ACI)~\citep{jimenez2023swebench,yang2024sweagentagentcomputerinterfacesenable}, and RepoAgent for repository-level documentation and maintenance~\citep{luo2024repoagent}.

\paragraph{GUI Agents}  
Research on GUI agents has advanced rapidly since the release of early grounding datasets such as ScreenSpot~\cite{cheng2024seeclick}.  
Within a short time, these datasets reached saturation, prompting a shift of focus from grounding individual elements to developing end-to-end agents capable of performing complete GUI-based tasks.  
Open-source efforts were the first to drive this transition, with systems such as CogAgent~\cite{hong2024cogagent}, OS-Atlas~\cite{wu2024atlas}, and Aguvis~\cite{xu2024aguvis}.  
This momentum was soon joined by industry initiatives, including OpenAI~\cite{openaioperator}, Anthropic~\cite{anthropic2025claude}, and Bytedance~\cite{qin2025ui}, leading to the rapid proliferation of computer-use agents.  
Early approaches were largely data-driven, relying on diverse human demonstrations for supervised fine-tuning.  
While these methods enabled the first generation of GUI agents, they suffered from limited generalization and poor robustness in complex environments.  
More recently, reinforcement learning (RL) has emerged as a promising direction, with systems such as ARPO~\cite{lu2025arpo} and Mobile-GUI-R1~\cite{shi2025mobilegui} applying RL-based training.

\paragraph{Game Agents}  
Game environments provide a natural testbed for studying interactive decision-making, where long-horizon control and strategic exploration are essential.  
Digital games have historically been central to AI research due to their complexity, diversity, and controllability.  
Seminal work ranges from classical board games such as Go~\citep{alphago}, to Atari benchmarks~\citep{atari}, to large-scale strategy games like StarCraft II~\citep{ma2024large}, and open-ended environments such as Minecraft~\citep{minedojo}.  
However, a key limitation of these efforts is their specificity: agents were typically optimized for a single game with tailored policies and parameters, hindering generalization across different environments~\citep{vpt,dqn,alphastar}.  

The emergence of LLMs and VLMs has shifted attention toward more generalist agents~\citep{reed2022gato}.  
Recent work explores their application to complex game scenarios, such as Pokémon~\citep{comanici2025gemini,claude37extendthink}.  
To cope with the long-horizon and multimodal nature of games, many approaches adopt workflow-style designs, equipping models with explicit modules for memory~\citep{wang2024jarvis,wang2023voyager} and planning~\citep{zhang2024proagent,shinn2023reflexion,yuan2023plan4mc}, or fine-tuning VLMs on specific titles for domain specialization~\citep{sima,li2025jarvis}.  
A key distinction lies in the interaction modality.  
Most existing systems depend on textual observations and predefined semantic actions exposed via game APIs~\citep{wang2023voyager,comanici2025gemini}.  
By contrast, our framework engages with games through the same modality used for GUI tasks—native GUI actions grounded in visual input.  
This unified interface integrates game environments with general computer-use scenarios, eliminating the need for additional handcrafted modules and highlighting the potential of building agents that generalize across diverse interactive digital settings.

\paragraph{Other Relevant Directions}
Beyond GUI and games, two other lines of research provide complementary insights into the design of interactive agents. Protocols such as MCP~\citep{Anthropic2024MCP} have introduced standardized mechanisms for flexible tool integration, enabling agents to seamlessly interact with search engines, file parsers, or external APIs. Building on this foundation, recent work has pursued two primary directions: end-to-end reinforcement learning~\cite{feng2025retoolreinforcementlearningstrategic,li2025torlscalingtoolintegratedrl,wang2025actingreasoningmoreteaching,song2025r1plus}, which directly optimizes multi-step reasoning with tool calls, and workflow-based methods~\cite{guo2024owllargelanguagemodel,qiu2025alitageneralistagentenabling}, which orchestrate tools through scripted procedures but often lack flexibility. Early studies mainly focused on simple tool-enhanced tasks such as HotpotQA or MathQA~\cite{li2025searcho1agenticsearchenhancedlarge,gou2024toratoolintegratedreasoningagent}, whereas more recent efforts have introduced harder benchmarks like BrowseComp~\cite{wei2025browsecomp}, where information is deliberately obfuscated across websites. These benchmarks reveal the limitations of models trained only on simple data~\cite{simpledeepsearcher} and have motivated research on synthesizing high-difficulty datasets~\cite{li2025websailornavigatingsuperhumanreasoning} and building multi-agent or planning-based systems~\cite{chen2024mindsearchmimickinghumanminds,huang2025manusearchdemocratizingdeepsearch}.  

In parallel, LLM-based code agents have reshaped software automation. SWE-bench~\citep{jimenez2023swebench} established a benchmark for repository-level issue resolution and inspired systems such as SWE-agent~\citep{yang2024sweagentagentcomputerinterfacesenable}. This was followed by richer datasets including SWE-Gym~\citep{pan2025trainingsoftwareengineeringagents}, SWE-Bench-Extra~\citep{badertdinov2024scaling}, SWE-ReBench~\citep{badertdinov2025swerebenchautomatedpipelinetask}, and Multi-SWE-RL~\citep{zan2025multiswebench}, which broaden task coverage and programming language diversity. Frameworks such as OpenHands~\citep{wang2025openhandsopenplatformai} advanced sandboxed agentic coding with execution feedback, while Terminal Bench~\citep{tbench_2025} emphasized command-line proficiency as a critical skill. Alongside, both proprietary models (Gemini~\citep{comanici2025gemini}, Claude~\citep{claude4intro}, GPT-5~\citep{gpt5intro}) and open-source efforts (Qwen-3 Coder~\citep{qwen3technicalreport}, Kimi-K2~\citep{kimiteam2025kimik2openagentic}, GLM-4.5~\citep{5team2025glm45agenticreasoningcoding}) have increasingly prioritized agentic coding capabilities. These trends illustrate how reinforcement learning, interactive feedback, and curated datasets are becoming central for scaling code agents.  
\section{Conclusion}
\label{sec:conclusion}
In this work, we presented UI-TARS-2, a native GUI-centered agent model designed to handle both structured computer-use tasks and dynamic, game-like interactive environments.  
The model is trained through an iterative pipeline that combines multi-turn reinforcement learning, supervised fine-tuning, rejection sampling, and continual pre-training, enabling continual improvement across heterogeneous domains.  
Our experiments show that while domain-specialized variants can achieve peak scores on individual benchmarks, UI-TARS-2 attains balanced and competitive performance across GUI, browser, mobile, and game tasks within a single unified system.  
Beyond benchmark results, our analysis (e.g., training dynamics and interaction scaling) yields practical insights into multi-turn agent RL.  
We also demonstrate that training on diverse environments promotes parameter sharing and capability transfer, giving rise to hybrid skills that integrate graphical interaction with more complex forms of reasoning and decision-making.  
Taken together, UI-TARS-2 represents a step toward more capable, reliable, and versatile computer-use agents, offering both empirical evidence and methodological principles to guide future research.

\clearpage

\bibliographystyle{plainnat}
\bibliography{main}

\clearpage
\section{Contributions}
\label{sec:contributions}
The authors are listed alphabetically by first name, with some names corresponding to internal aliases used within the company.
\setlength{\parskip}{0pt} 
\setlength{\itemsep}{0pt} 
\setlength{\parsep}{0pt}  
\begin{multicols}{2}

\subsection*{Algorithm}

\subsubsection*{Core Contributors}
Haoming Wang \\
Haoyang Zou \\
Huatong Song \\
Jiazhan Feng \\
Junjie Fang \\
Junting Lu \\
Longxiang Liu \\
Qinyu Luo \\
Shihao Liang \\
Shijue Huang \\
Wanjun Zhong \\
Yining Ye \\
Yujia Qin \\
Yuwen Xiong \\
Yuxin Song \\
Zhiyong Wu

\subsubsection*{Contributors}
Aoyan Li \\
Bo Li \\
Chen Dun \\
Chong Liu \\
Daoguang Zan \\
Fuxing Leng \\
Hanbin Wang \\
Hao Yu \\
Haobin Chen \\
Hongyi Guo \\
Jing Su \\
Jingjia Huang \\
Kai Shen \\
Kaiyu Shi \\
Lin Yan \\
Peiyao Zhao \\
Pengfei Liu \\
Qinghao Ye \\
Renjie Zheng \\
Shulin Xin \\
Wayne Xin Zhao \\
Wen Heng \\
Wenhao Huang \\
Wenqian Wang \\
Xiaobo Qin \\
Yi Lin \\
Youbin Wu \\
Zehui Chen \\
Zihao Wang

\subsection*{Infra}
\subsubsection*{Core Contributors}
Baoquan Zhong \\
Xinchun Zhang \\
Xujing Li \\
Yuanfan Li \\
Zhongkai Zhao

\subsubsection*{Contributors}
Chengquan Jiang \\
Faming Wu \\
Haotian Zhou \\
Jinlin Pang \\
Li Han \\
Qi Liu \\
Qianli Ma \\
Siyao Liu \\
Songhua Cai \\
Wenqi Fu \\
Xin Liu \\
Yaohui Wang \\
Zhi Zhang

\subsection*{Data}
\subsubsection*{Core Contributors}
Bo Zhou \\
Guoliang Li \\
Jiajun Shi \\
Jiale Yang \\
Jie Tang \\
Li Li \\
Qihua Han \\
Taoran Lu \\
Woyu Lin \\
Xiaokang Tong \\
Xinyao Li \\
Yichi Zhang \\
Yu Miao \\
Zhengxuan Jiang \\
Zili Li \\
Ziyuan Zhao

\subsubsection*{Contributors}
Chenxin Li \\
Dehua Ma \\
Feng Lin \\
Ge Zhang \\
Haihua Yang \\
Hangyu Guo \\
Hongda Zhu \\
Jiaheng Liu \\
Junda Du \\
Kai Cai \\
Kuanye Li \\
Lichen Yuan \\
Meilan Han \\
Minchao Wang \\
Shuyue Guo \\
Tianhao Cheng \\
Xiaobo Ma \\
Xiaojun Xiao \\
Xiaolong Huang \\
Xinjie Chen \\
Yidi Du \\
Yilin Chen \\
Yiwen Wang \\
Zhaojian Li \\
Zhenzhu Yang \\
Zhiyuan Zeng

\subsection*{Application}
Chaolin Jin \\
Chen Li \\
Hao Chen \\
Haoli Chen \\
Jian Chen \\
Qinghao Zhao

\subsection*{Supervisor}
Guang Shi

\end{multicols}

\clearpage


\end{document}